\documentclass[lettersize,journal]{IEEEtran}
\usepackage{amsmath,amsfonts}
\usepackage{amsthm}
\usepackage{algorithmic}
\usepackage{algorithm}
\usepackage{array}
\usepackage[caption=false,font=normalsize,labelfont=sf,textfont=sf]{subfig}
\usepackage{textcomp}
\usepackage{stfloats}
\usepackage{url}
\usepackage{verbatim}
\usepackage{graphicx}
\usepackage{xcolor}
\usepackage{cite}
\usepackage{hyperref}
\usepackage{nicematrix}
\usepackage{xcolor}
\usepackage{colortbl}  
\usepackage{booktabs} 
\usepackage{multirow} 
\usepackage{adjustbox} 

\definecolor{baselinecolor}{gray}{0.9}
\definecolor{deltacolor}{HTML}{E0F7FA}

\newcommand{\updated}[1]{\textcolor{black}{#1}}

\hypersetup{
	colorlinks=true,       
	linkcolor=blue,          
	citecolor=green,        
	filecolor=magenta,      
}

\def\R{\mathrm {\mathbf{R}}}
\def\RR{\mathbb R}

\def\W{\mathrm {\mathbf{W}}}
\def\S{\mathrm {\mathbf{S}}}
\def\Z{\mathrm {\mathbf{Z}}}

\newtheorem{theorem}{Theorem}
\newtheorem{lemma}{Lemma}
\usepackage{stmaryrd}

\usepackage{setspace}
\begin{document}

\title{Embedding-Driven Data Distillation for 360-Degree IQA With Residual-Aware Refinement}

\author{Abderrezzaq Sendjasni, ~\IEEEmembership{Member,~IEEE,}
        Seif-Eddine Benkabou, 
        Mohamed-Chaker Larabi, ~\IEEEmembership{Senior Member,~IEEE}
\thanks{This work is partially funded by the Nouvelle-Aquitaine Research Council under project REALISME AAPR2022-2021-17027310. \\
}
}
\markboth{Journal of \LaTeX\ Class Files,~Vol.~14, No.~8, August~2021}%
{Abderrezzaq \MakeLowercase{\textit{et al.}}: A Two-Fold Patch Selection Approach for Improved 360-Degree Image Quality Assessment}

\maketitle

\begin{abstract}

\updated{This article identifies and addresses a fundamental bottleneck in data-driven 360-degree image quality assessment (IQA): the lack of intelligent, sample-level data selection. Hence, we propose a novel framework that introduces a critical refinement step between patches sampling and model training. The core of our contribution is an embedding similarity-based selection algorithm that distills an initial, potentially redundant set of patches into a compact, maximally informative subset. This is formulated as a regularized optimization problem that preserves intrinsic perceptual relationships in a low-dimensional space, using residual analysis to explicitly filter out irrelevant or redundant samples. Extensive experiments on three benchmark datasets (CVIQ, OIQA, MVAQD) demonstrate that our selection enables a baseline model to match or exceed the performance of using all sampled data while keeping only 40-50\% of patches. Particularly, we demonstrate the universal applicability of our approach by integrating it with several state-of-the-art IQA models, incleasy to deploy. Most significantly, its value as a generic,uding CNN- and transformer-based architectures, consistently enabling them to maintain or improve performance with 20-40\% reduced computational load. This work establishes that adaptive, post-sampling data refinement is a powerful and widely applicable strategy for achieving efficient and robust 360-degree IQA. 
}

\end{abstract}


\section{Introduction}
\label{sec:intro}

\IEEEPARstart{T}{he} proliferation of virtual reality (VR), augmented reality (AR), and immersive multimedia has created an unprecedented demand for high-quality 360-degree images and videos~\cite{ALAIN20233, 360_growth}. Unlike conventional 2D media, these spherical formats offer an interactive, immersive experience, making the assurance of their visual fidelity paramount to user engagement and technological adoption~\cite{perkis2020qualinet, ALAIN20233}. However, the very properties that enable immersion, the spherical geometry and interactive viewport-dependent navigation, also make perceptual quality assessment exceptionally challenging. Traditional 2D image quality assessment (IQA) metrics, when applied directly, struggle with projection-induced distortions and fail to account for the complex interplay of human visual exploration and spherical content~\cite{8434258}.

Deep learning has emerged as a powerful paradigm to address these challenges, learning intricate mappings from visual data to human perceptual scores. Subsequent models have demonstrated significant improvements over conventional approaches like PSNR and SSIM~\cite{Sun2017, 8486584}. A common strategy among these learning-based methods is content sampling, where 360-degree images are represented as a collection of viewports~\cite{8953510, 9163077, 9964240}, cubemap faces~\cite{8702664, 9334423}, or patches~\cite{8638985, s23218676, 9428390}. The selection of these samples is often guided by heuristics such as uniform projection sampling, latitude-based importance, or predicted visual scanpaths~\cite{yang2022tvformer}. While these strategies ensure coverage of the spherical surface, they operate under a critical and often unstated assumption: \textbf{that all sampled regions are equally valuable for training a quality regression model}. In reality, these methods can oversample geometrically distorted or perceptually irrelevant areas, introduce biases from saliency or scanpath predictors, and create datasets with high redundancy. This not only imposes unnecessary computational burdens but can also limit model generalization by diluting the training signal with uninformative or noisy samples.

This work identifies and addresses a fundamental bottleneck in data-driven 360-IQA: \textbf{the lack of intelligent, post-sampling selection}. While existing research meticulously decides \textit{"where"} to sample from, it largely neglects the crucial step of determining \textit{"which"} of the sampled patches are most informative for the quality prediction task. Consequently, models are trained on large, redundant datasets where a significant portion of the data may contribute little to learning. This problem of sample redundancy and noise is well-recognized in general machine learning, where instance selection techniques are used to curate compact, high-quality training sets~\cite{garcia2015data, kuncheva2019instance}. However, their application to perceptual IQA, particularly in the 360-degree domain, remains largely unexplored.

\updated{To bridge this gap, we introduce a novel and agnostic patch selection framework not designed as a new quality metric, but as a general-purpose strategy to enhance the data efficiency and robustness of existing 360-IQA pipelines. Our approach is built on a key hypothesis: \textbf{that the most informative patches for quality assessment are those whose perceptual characteristics are well-preserved in a structured, low-dimensional embedding space}. The proposed framework is fundamentally agnostic to the initial source of visual patches, allowing it to be seamlessly integrated with any existing sampling strategy (projection- or perception-aware patches sampling). Its core contribution is a novel optimization problem that selects a final, informative subset of patches by preserving the intrinsic similarity structure of their embeddings. A key component of this formulation is a residual analysis that explicitly identifies and filters out redundant samples, ensuring a concise and representative patch set for accurate quality assessment.}

\updated{This mechanism ensures that the final selected patch set is not only compact but also maximally informative for quality assessment. We rigorously evaluate our method not by competing against specific SOTA metrics, but by demonstrating its value and resilience as a universal preprocessing step across different initial sampling strategies and datasets. Extensive experiments on CVIQ, OIQA, and MVAQD benchmarks show that our selection framework enables models to match or exceed baseline performance using only 40-50\% of the data, and can be seamlessly integrated to improve various SOTA architectures.}

\updated{The key contributions of this work are summarized as follows:
\begin{itemize}
    \item We propose a novel agnostic patch selection framework that enhances the data efficiency and robustness of 360-degree IQA models. It operates as a universal preprocessing step by refining an existing set of patches through intelligent, embedding-based refinement.
    \item We introduce a core embedding similarity-based selection algorithm, formulated as a regularized optimization problem with residual analysis, to automatically identify and retain the most perceptually relevant patches from any initial sample pool.
    \item We provide extensive empirical and statistical validation, demonstrating consistent performance gains and computational savings across multiple datasets and, crucially, across multiple initial sampling strategies. Furthermore, we show the general applicability of our method by successfully integrating it with several state-of-the-art IQA models.
\end{itemize}
}

The remainder of the paper is organized as follows. Section~\ref{sec:related_work} provides a comprehensive review of related work. Section~\ref{sec:meth} details the proposed methodology. Section~\ref{sec:discussion} presents a thorough experimental validation and discussion. Finally, conclusions and future work are outlined in Section~\ref{sec:conc}.

\section{Related Work}
\label{sec:related_work}

This section reviews the evolution of 360-degree IQA, tracing the progression from traditional 2D metric adaptations to sophisticated deep learning models. We focus particularly on how these methods represent and select visual content from the spherical representation of the content, which directly contextualizes the contribution of our work.

\subsection{From 2D Adaptations to Deep Learning}

Early work in 360-degree IQA focused on adapting established 2D metrics, such as PSNR and SSIM, to handle the spherical-to-planar projection inherent in formats like the Equirectangular projection (ERP)~\cite{Sun2017, 8486584}. These methods accounted for geometric distortions but largely failed to incorporate the perceptual nuances of immersive viewing and user interaction, leading to limited correlation with human opinion scores. This limitation catalyzed the shift towards data-driven, deep learning-based approaches.

Deep learning models, particularly CNNs, offered a paradigm shift. A dominant strategy emerged using multi-channel architectures, where a 360-degree image is decomposed into multiple planar representations, such as viewports~\cite{9163077, 9506044, fan2024omiqnet}, cubemap faces~\cite{8702664, 9432940, zhou2023perception, 10297422}, or patches~\cite{8638985}, which are then processed by parallel, often pre-trained, networks. While these methods showed significant improvements over traditional metrics, they introduced critical challenges: (1) computational complexity, as models scaled to use six~\cite{9163077}, twenty~\cite{8702664}, or even thirty-two~\cite{8638985} sub-networks; and (2) representational rigidity, where the fixed, predefined sampling strategy, such as a specific cube-map layout, may not adapt well to diverse content or distortion types. The effectiveness of these models became heavily dependent on the chosen input representation~\cite{9791414}, highlighting a fundamental reliance on the initial data selection step.

\subsection{Incorporating Perceptual Cues and Advanced Architectures}

To better align with human perception, subsequent research has increasingly integrated user behavior models and advanced architectures. This includes leveraging the equator bias~\cite{yan2025omnidirectional} to select viewports around this region of ERP image known to attract the visual attention, visual saliency to weight regions based on their likelihood of being fixated upon~\cite{10096750, tofighi2024omnidirectional}, and modeling visual scanpaths to simulate dynamic user exploration~\cite{NEURIPS2023_ccf4a732, sendjasni2023pw}.

A significant trend involves using transformer architectures to capture long-range dependencies in spherical content. Notably, TVFormer~\cite{yang2022tvformer} uses predicted head trajectories to sample viewports, employing a hybrid CNN-Transformer architecture to capture both global and local spatio-temporal features. A limitation of this approach is its reliance on a single trajectory per image. Assessor360~\cite{NEURIPS2023_ccf4a732} addresses this by generating multiple exploration sequences, though its sampling remains constrained by the initial viewpoint. Other works have focused on hybrid architectures and multi-scale feature fusion. ST360IQ~\cite{10096750} and the model by Liu et al.~\cite{liu2023dual} combine CNNs and vision transformers (ViTs) to fuse multi-scale features from saliency-guided regions or cubemap faces. More recently, Max360IQ~\cite{yan2025max360iq} has advanced this trend with multi-axis attention and semantic-guided regression.

\updated{A common thread among these perceptually-guided and architecturally advanced models is their sophisticated focus on \textit{"where to look"}, they employ complex mechanisms to identify potentially important regions. However, they typically use \textit{"all sampled data"}, such as all viewports along a scanpath or the six cubemap faces, for training and inference. This approach overlooks the inherent data correlation and redundancy within the constructed training sets, where multiple similar or uninformative samples may dilute the learning signal. Consequently, while these methods excel at regional identification, they may underuse the most informative samples and remain susceptible to overfitting on their specific sampling strategy, ultimately limiting their generalization potential.}

\subsection{Data Selection for Efficient and Robust IQA}

A critical yet under-explored issue for improving 360-degree IQA is adaptive data selection at the sample level. In general machine learning, instance selection and core-set methods are well-established for creating compact, representative training subsets by removing redundant and noisy samples, thereby improving model robustness, efficiency, and scalability~\cite{kuncheva2019instance, chen2022deep}.



\updated{While current 360-IQA methods meticulously select regions to sample from, such as via saliency or scanpaths, they largely neglect sample-level curation thereafter. This results in training sets with high internal redundancy. The application of adaptive selection strategies to refine these initial samples and identify the most informative subset for the quality prediction task remains a significant gap in the literature. Hence, we propose an embedding-similarity-driven strategy to fill this gap, which is detailed in the following section.}

\section{The Proposed Methodology}
\label{sec:meth}

\begin{figure*}[ttbp]
    \centering
    \includegraphics[width=\textwidth]{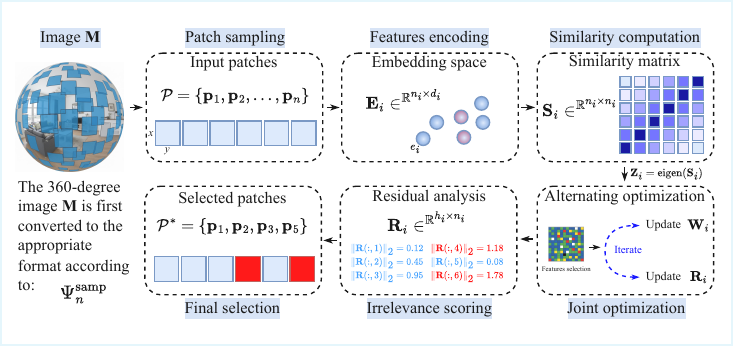}
    \caption{\updated{Overview of the proposed embedding similarity-based selection framework. The algorithm transforms patch embeddings to a low-dimensional space while preserving similarity structure, then uses residual analysis to identify and select the most informative patches.}}
    \label{fig:fram}
\end{figure*}

\updated{This section details the core contribution of our work: a novel embedding similarity-based selection framework designed to enhance the data efficiency of 360-degree IQA models. The overall pipeline, illustrated in Fig.~\ref{fig:fram}, operates through a sequence of steps to solve a fundamental data curation problem. It begins with (1) a flexible visual patch identification from the input 360-degree image, followed by (2) patch sampling to generate a comprehensive initial set. These patches are then (3) encoded into a feature space to generate dense embeddings. The core of our method lies in the subsequent analysis of these embeddings: (4) similarity computation constructs a relational graph of the patches, which feeds into (5) a joint optimization process that performs simultaneous feature selection and redundancy detection. This optimization yields (6) a residual matrix used for irrelevance scoring and ranking of the patches. The final output is (7) a curated subset of patches, selected based on this ranking, which is then used for quality regression. This end-to-end process ensures the model trains on a compact, maximally informative dataset, leading to more robust and efficient quality assessment.}

\subsection{Notations}
We first define the mathematical notations used throughout this article. Sets are denoted by calligraphic letters (such as $\mathcal{D}$), matrices by bold uppercase (such as $\mathbf{X}$), vectors by bold lowercase (such as $\mathbf{a}$), and scalars by regular letters (such as $n$). A comprehensive summary is provided in Table~\ref{Symbols}.

\begin{table}[htbp]
	\centering
	\caption{Summary of key symbols and notations.}
	\begin{tabular}{p{1.9cm}p{6.1cm}}
		\toprule
		\textbf{Symbol} & \textbf{Definition}\\
		\midrule
        $\mathcal{D}$ & Collection of $m$ source images.\\
        $\mathbf{M}_i$ & The $i$-th 360° image.\\
        $\mathcal{P}$ & Collection of $n$ patches $\mathbf{p}_i \in \mathbb{R}^{x \times y \times 3}$.\\
        $\mathbf{E}_i$ & Embedding matrix for image $\mathbf{M}_i$, where each row is a patch embedding $\mathbf{e}_j \in \mathbb{R}^{d_i}$.\\
        $n_i$ & Number of patches sampled from image $\mathbf{M}_i$.\\
        $d_i$ & Dimensionality of a patch embedding (2048 for ResNet-50).\\
        $h$ & Dimension of the low-dimensional projection space ($h \ll d_i$).\\
        $\mathbf{W}_i \in \mathbb{R}^{d_i \times h}$ & Transformation matrix for image $\mathbf{M}_i$.\\
        $\mathbf{R}_i \in \mathbb{R}^{h \times n_i}$ & Residual matrix for image $\mathbf{M}_i$.\\
        $\mathbf{I}_k$ & Identity matrix of size $k \times k$.\\
        $\| \cdot \|_{2,1}$ & $\ell_{2,1}$ norm of a matrix (sum of $\ell_2$ norms of rows).\\
        $\| \cdot \|_{F}$ & Frobenius norm of a matrix.\\
        $\mathcal{Q}$ & Set of predicted quality scores.\\
		\bottomrule
	\end{tabular}
	\label{Symbols}
\end{table}


\subsection{Visual Patch Sampling}
\label{sec:vis_patch_selec}

\updated{
The high resolution (typically 4K and beyond) and spherical nature of 360-degree images make their entire processing computationally prohibitive and perceptually suboptimal~\cite{9791414}. Consequently, a critical first step in any patch-based 360-IQA pipeline is to sample a finite set of representative patches from the original spherical content.}

\updated{
Let $\mathbf{M} \in \mathbb{R}^{H \times W \times 3}$ represent an input 360-degree image in its equirectangular projection. A sampling method $\Psi$ is defined as a function that extracts $n$ patches from $\mathbf{M}$:}

\updated{
\begin{equation}
    \Psi_{n}^{\mathrm{samp}} : \mathbf{M} \rightarrow \mathcal{P} \subset \mathbb{R}^{n \times x \times y \times 3},
\end{equation}}
\updated{where $\mathrm{samp}$ denotes a specific sampling strategy, and $\mathcal{P} = \{\mathbf{p}_1, \mathbf{p}_2, \dots, \mathbf{p}_n\}$ is the resulting set of $n$ patches, each of size $x \times y$ pixels.}

\updated{
Various sampling strategies exist, ranging from uniform grid sampling on the ERP to perceptually-driven approaches based on spherical geometry, saliency, or predicted scanpaths. The choice of $\Psi_{n}^{\mathrm{samp}}$ introduces an initial, implicit bias by determining which regions of the spherical image are represented. For instance, a strategy might prioritize the equatorial regions due to their higher likelihood of being viewed, or it might focus on areas of high visual saliency.}

\updated{
In this work, we treat this sampling step as a modular and necessary precursor. Its primary purpose is to generate an initial, and potentially redundant, set of candidate patches $\mathcal{P}$ that comprehensively cover the visual content. The specific implementation of $\Psi$ can be chosen based on the application's needs; our core contribution is agnostic to this choice. The key insight is that regardless of the sampling strategy, the resulting set $\mathcal{P}$ is unlikely to be optimally concise or informative for the quality regression task. This initial sampling sets the stage for the crucial, novel step that follows: the embedding similarity-based selection designed to distill $\mathcal{P}$ into a maximally informative subset.}

\subsection{Patch Encoding and Feature Representation}
\label{sec:patch_encoding}

The sampled patches $\mathcal{P}$ contain high-dimensional pixel data. To obtain a more robust and compact representation for analysis, we encode them into a latent feature space using a CNN. This transformation captures semantically meaningful patterns essential for various image processing tasks, including quality assessment.

Let $f_{\text{net}}(\cdot)$ represent a CNN encoder. For a set of patches $\mathcal{P}$, it produces a set of feature maps. To condense these spatial feature maps into a single vector per patch, we apply global average pooling (GAP), denoted $g(\cdot)$. This results in a final embedding matrix:
\begin{equation}\label{eq:enc}
\mathbf{E}_i = g(f_{\text{net}}(\mathcal{P})) = g(f_{\text{net}}(\Psi_{n}^{\mathrm{samp}}(\mathbf{M}_i))).
\end{equation}

Here, $\mathbf{E}_i \in \mathbb{R}^{n \times c}$ is the matrix of patch embeddings, where $c$ is the dimensionality of the output feature vector.

We use a ResNet-50 model~\cite{resnet}, pre-trained on ImageNet, as our encoder $f_{\text{net}}$. ResNet's residual connections facilitate the training of deep networks, enabling the extraction of rich, hierarchical features. The resulting embeddings ($c=2048$) encapsulate essential visual characteristics, such as texture, structure, and color, forming a powerful basis for the subsequent selection and regression tasks.

While the initial sampling (Sec.~\ref{sec:vis_patch_selec}) ensures diversity, the resulting set $\mathbf{E}_i$ may still contain redundancies or uninformative elements. The next section details our selection algorithm based on embedding similarity that refines this set to retain only the most informative patches for quality prediction. This represents the core contribution of this work.

\subsection{Embedding Similarity-Based Selection}
\label{sec:emb_sel}

\updated{For the proposed algorithm, the goal is to identify and retain the most informative embeddings while discarding those that are redundant or irrelevant for the quality prediction task.} We achieve this by formulating a novel optimization problem that jointly preserves the intrinsic similarity structure of the embeddings and identifies outliers through residual analysis.

\updated{\subsubsection{Problem Formulation}
Let $\mathcal{D} = \{\mathbf{M}_1, \dots, \mathbf{M}_m\}$ be a collection of $m$ images, where each image $\mathbf{M}_i$ is represented by a set of $n_i$ patch embeddings, $\mathbf{E}_i = [\mathbf{e}_1, \dots, \mathbf{e}_{n_i}]\in \mathbb{R}^{n_i \times d_i}$. Our objective is to select a subset of these patches that best represents the perceptual quality of the original image.}

The fundamental assumption is that perceptually similar patches cluster together in the embedding space. A high-quality subset should preserve this underlying similarity structure. To formalize this, we seek a transformation matrix $\mathbf{W}_i \in \mathbb{R}^{d_i \times h}$ that maps the high-dimensional embeddings $\mathbf{E}_i$ to a lower-dimensional space $\mathbf{E}_i\mathbf{W}_i$ ($h \ll d_i$) while preserving the pairwise similarities~\cite{similarity}. This can be formulated as minimizing the discrepancy between the similarity matrices in the original and reduced spaces:
\begin{equation}
\label{eq:projection_problem}
\min_{\mathbf{W}_i} \left\| (\mathbf{E}_i\mathbf{W}_i)(\mathbf{E}_i\mathbf{W}_i)^\top - \mathbf{S}_i \right\|^{2}_{F} + \alpha \| \mathbf{W}_i \|_{2,1},
\end{equation}
where:
\begin{itemize}
    \item $\mathbf{S}_i \in \mathbb{R}^{n_i \times n_i}$ is the similarity matrix computed from $\mathbf{E}_i$. We investigate three distance metrics to construct $\mathbf{S}_i$: Euclidean (EUC) for straight-line distance, Manhattan (MAN) for robustness to outliers, and Mahalanobis (MAH) to account for feature correlations~\cite{mahalanobis2cv, bishop2006pattern}.
    \item The $\ell_{2,1}$-norm regularization term $\alpha \| \mathbf{W}_i \|_{2,1}$ promotes sparsity in $\mathbf{W}_i$, ensuring that the transformation relies on only the most discriminative features.
\end{itemize}

\subsubsection{Tractable Reformulation}
Directly solving Eq.~\eqref{eq:projection_problem} is NP-hard. However, the optimal solution to the relaxed problem of finding a low-rank approximation of the similarity matrix,
\begin{equation}
\label{relaxation_problem}
    \min_{\mathbf{X} \in \mathbb{R}^{n\times h}} \| \mathbf{X} \mathbf{X}^\top - \mathbf{S}_i \|^{2}_{F},
\end{equation}
is given by the eigen-decomposition of $\mathbf{S}_i$. Specifically, the solution is $\mathbf{X}^* = \mathbf{\Lambda}_{n\times h} \sqrt{\mathbf{D}_h}$, where $\mathbf{\Lambda}$ contains the top-$h$ eigenvectors and $\mathbf{D}_h$ is the diagonal matrix of the corresponding largest eigenvalues. Thus, we define $\mathbf{Z}_i = \mathbf{X}^*$ such that $\mathbf{S}_i = \mathbf{Z}_i \mathbf{Z}_i^\top$.

This allows us to reformulate the intractable problem in Eq.~\eqref{eq:projection_problem} into a tractable one: finding a transformation $\mathbf{W}_i$ such that $\mathbf{E}_i\mathbf{W}_i$ approximates this optimal low-dimensional representation $\mathbf{Z}_i$:
\begin{equation}
\label{eq:reduced_problem}
\min_{\mathbf{W}_i} \| \mathbf{E}_i\mathbf{W}_i - \mathbf{Z}_i \|^{2}_{F} + \alpha \| \mathbf{W}_i \|_{2,1}.
\end{equation}

\subsubsection{Residual Analysis for Patch Selection}
The reformulated problem in Eq.~\eqref{eq:reduced_problem} learns a transformed space but does not explicitly select patches. To address this, we introduce a \textit{residual matrix} $\mathbf{R}_i \in \mathbb{R}^{h \times n_i}$ into the objective. The key insight is that if the similarity structure of a patch embedding $\mathbf{e}_j$ is well-preserved after transformation, the residual for that patch will be small. Conversely, a large residual indicates an irrelevant patch.

\updated{We define the residual as the deviation between the transformed embeddings and the optimal representation, accounting $\boldsymbol{\Theta}$ assumed to be multidimentional normal distribution~\cite{residual_analysis}:}
\begin{equation}
    \label{residual_matrix}
    \mathbf{R}_i = (\mathbf{E}_i\mathbf{W}_i)^\top - \mathbf{Z}_i^\top - \boldsymbol{\Theta}.
\end{equation}
By promoting sparsity in $\mathbf{R}_i$, we force the model to explain the data primarily through the transformation $\mathbf{W}_i$, using large residuals only for patches that cannot be fit well, these are the candidates for removal.

Integrating this into our framework yields the final joint optimization problem:
\begin{equation}  \label{objectifve}
\min_{\mathbf{W}_i,\mathbf{R}_i} \| \mathbf{E}_i\mathbf{W}_i - \mathbf{Z}_i - \mathbf{R}_{i}^\top \|_F^2 + \alpha \| \mathbf{W}_i\|_{2,1} + \beta \| \mathbf{R}_i \|_{2,1}.
\end{equation}
Here, $\beta$ is a hyperparameter that controls the sparsity of $\mathbf{R}_i$, effectively determining the number of patches selected. A larger $\beta$ results in a smaller subset of retained patches.

\subsection{Optimization Algorithm}
\label{sec:optimization}
The objective function in Eq.~\eqref{objectifve} is non-convex jointly in $\mathbf{W}_i$ and $\mathbf{R}_i$ but convex in each variable separately. We therefore employ an alternating optimization strategy. Furthermore, the $\ell_{2,1}$-norms are non-smooth, so we use a common relaxation technique~\cite{l21-norms-paper}, approximating $\|\mathbf{M}\|_{2,1}$ by $\text{Tr}(\mathbf{M}^\top \mathbf{D}_{\mathbf{M}} \mathbf{M})$, where $\mathbf{D}_{\mathbf{M}}$ is a diagonal matrix with elements $d_{jj} = 1 / (2 \|\mathbf{m}_{j:}\|_2)$.

The resulting alternating minimization steps are as follows:

\textbf{Step 1: Update the Transformation Matrix $\mathbf{W}_i$ (with $\mathbf{R}_i$ fixed).}
The Lagrangian for this sub-problem is:
\begin{equation}
\label{Lagrangian_W_i}
\mathcal{L}(\mathbf{W}_i) = \text{Tr}\left( \mathbf{W}_i^\top \mathbf{E}_i^\top \mathbf{E}_i \mathbf{W}_i - 2 \mathbf{W}_i^\top \mathbf{E}_i^\top (\mathbf{R}_i^\top + \mathbf{Z}_i) \right) + \alpha \| \mathbf{W}_i \|_{2,1}.
\end{equation}
Taking the derivative with respect to $\mathbf{W}_i$ and setting it to zero yields the closed-form update:
\begin{equation}
\label{W_i_solution}
\mathbf{W}_i = \left( \mathbf{E}_i^\top \mathbf{E}_i + \alpha \mathbf{D}_{\mathbf{W}_i} \right)^{-1} \mathbf{E}_i^\top (\mathbf{R}_i^\top + \mathbf{Z}_i).
\end{equation}

\textbf{Step 2: Update the Residual Matrix $\mathbf{R}_i$ (with $\mathbf{W}_i$ fixed).}
The Lagrangian for this sub-problem is:
\begin{equation}
\label{Lagrangian_R_i}
\mathcal{L}(\mathbf{R}_i) = \text{Tr}\left( \mathbf{R}_i^\top \mathbf{R}_i - 2 \mathbf{R}_i^\top (\mathbf{E}_i \mathbf{W}_i - \mathbf{Z}_i) \right) + \beta \| \mathbf{R}_i \|_{2,1}.
\end{equation}
Similarly, setting the derivative to zero gives the update rule for $\mathbf{R}_i$:
\begin{equation}
\label{R_i_solution}
\mathbf{R}_i = (\mathbf{E}_i \mathbf{W}_i - \mathbf{Z}_i)^\top \left( \mathbf{I}_{n_i} + \beta \mathbf{D}_{\mathbf{R}_i} \right)^{-1}.
\end{equation}

We iteratively update $\mathbf{W}_i$ and $\mathbf{R}_i$ until convergence. The complete procedure is summarized in Algorithm \ref{algo1}. After convergence, the $\ell_2$-norm of each column $\mathbf{R}_i(:,j)$ represents the \textit{irrelevance score} of the $j$-th patch. Patches are sorted by this score in ascending order, and the top-$k$ most relevant patches are selected for the final quality assessment model.

\begin{algorithm}
\caption{\textbf{Embeddings selection for $\mathbf{M}_i$}}
\label{algo1}
\begin{algorithmic}[1]
    \STATE {\bfseries input:} {$\mathbf{E_i}$, $\alpha$, $\beta$ and $h$}.
    \STATE {\bfseries output:} {Top $k$ relevant embeddings $\textbf{e}_j$ for $\textbf{E}$.}
    \STATE Compute the similarity matrix  of $\S$ embeddings $\textbf{E}$.%
     \STATE Get $\Z$ such as $\S_i=\Z\Z^\top$, ($\Z \in \RR^{n\times h})$.
     \STATE Initialize the residual matrix $\R$ by $\mathbf{[0]}_{h\times n}$
       \WHILE{convergence not reached}
        \STATE Update $\W_i$ and  $\R_i$   by Eq. \eqref{W_i_solution} and Eq. \eqref{R_i_solution} respectively.
        \ENDWHILE
        \STATE Rank the embedding $\textbf{e}_j$ of $\mathbf{E_i}$ according to the $\ell_2$-norm of $\R_i$ columns in ascending order $(j=1,\dots, n)$.
\end{algorithmic}
\end{algorithm}

\subsection{Complexity Analysis}

To assess the scalability of our proposed selection algorithm (Algorithm~\ref{algo1}), we analyze its computational complexity per image.

\begin{lemma}
The per-iteration computational complexity of Algorithm~\ref{algo1} for a single image $\mathbf{M}_i$ is $\mathcal{O}(h d_i^2 + h d_i n_i)$, where $d_i$ is the embedding dimension, $n_i$ is the number of patches, $h$ is the reduced dimension, and $t$ is the number of iterations. The overall complexity is $\mathcal{O}(t \cdot h \cdot d_i (d_i + n_i))$.
\end{lemma}

\begin{proof}
The complexity is dominated by the two update steps in each iteration:
\begin{enumerate}
    \item \textbf{Updating $\mathbf{W}_i$ (Eq.~\ref{W_i_solution}):} Solving the linear system $\left(\mathbf{E}_i^\top \mathbf{E}_i + \alpha \mathbf{D}_{\mathbf{W}_i} \right) \mathbf{W}_i = \mathbf{E}_i^\top (\mathbf{R}_i^\top + \mathbf{Z}_i)$ involves a matrix inversion of size $d_i \times d_i$, which has a complexity of $\mathcal{O}(d_i^3)$. However, since we solve for $h$ columns independently and the matrix $\mathbf{E}_i^\top \mathbf{E}_i$ is fixed per image, efficient solvers can achieve a complexity of $\mathcal{O}(h d_i^2)$.

    \item \textbf{Updating $\mathbf{R}_i$ (Eq.~\ref{R_i_solution}):} This matrix multiplication and scaling operation requires $\mathcal{O}(h d_i n_i)$ time, as it involves matrices of dimensions $(n_i \times d_i) \times (d_i \times h)$ and subsequent scaling of an $n_i \times h$ matrix.
\end{enumerate}
Summing the per-iteration costs and multiplying by the number of iterations $t$ yields the overall complexity of $\mathcal{O}(t \cdot h \cdot d_i (d_i + n_i))$. In practice, $h \ll d_i$ and the algorithm converges quickly ($t$ is small), making it efficient for the patch selection task.
\end{proof}

\subsection{Convergence Analysis}
\label{sec:conv}

The alternating minimization strategy employed in Algorithm~\ref{algo1} is guaranteed to converge.

\begin{lemma}\label{lemma:inequality}
For any two non-zero vectors $\mathbf{a}, \mathbf{b} \in \mathbb{R}^p$, the following inequality holds:
\begin{equation}
    \|\mathbf{a}\|_2 - \frac{\|\mathbf{a}\|_2^2}{2\|\mathbf{b}\|_2} \le \|\mathbf{b}\|_2 - \frac{\|\mathbf{b}\|_2^2}{2\|\mathbf{b}\|_2}.
\end{equation}
\end{lemma}
\begin{proof}
This is a direct consequence of the fact that the function $f(x) = \sqrt{x}$ is concave for $x > 0$. The inequality follows from the properties of the first-order Taylor expansion~\cite{l21-norms-paper}.
\end{proof}

\begin{theorem}
The sequence $\{ \Phi(\mathbf{W}_i^{(h)}, \mathbf{R}_i^{(h)}) \}_{i=1}^{50}$ generated by Algorithm~\ref{algo1} is non-increasing, i.e., 
$\Phi(\mathbf{W}_i^{(h+1)}, \mathbf{R}_i^{(h+1)}) \le \Phi(\mathbf{W}_i^{(h)}, \mathbf{R}_i^{(h)})$ for all $h$, and thus the algorithm converges.
\end{theorem}

\begin{proof}
The proof establishes that each update step decreases the objective function.

\textbf{Part 1: Update of $\mathbf{W}_i$ (with $\mathbf{R}_i$ fixed).}
Let $\mathbf{R}_i$ be fixed at iteration $h$. The update for $\mathbf{W}_i^{(h+1)}$ is the solution to:
\begin{equation}
    \mathbf{W}_i^{(h+1)} = \arg\min_{\mathbf{W}_i} \| \mathbf{E}_i\mathbf{W}_i - \mathbf{Z}_i - (\mathbf{R}_i^{(h)})^\top \|_F^2 + \alpha \| \mathbf{W}_i \|_{2,1}.
\end{equation}
Using the relaxation $\| \mathbf{W}_i \|_{2,1} \approx \text{Tr}(\mathbf{W}_i^\top \mathbf{D}_{\mathbf{W}} \mathbf{W}_i)$ with $\mathbf{D}_{\mathbf{W}}$ defined from $\mathbf{W}_i^{(h)}$, we have the following inequality at the optimum:
\begin{equation}
\begin{aligned}
&\| \mathbf{E}_i\mathbf{W}_i^{(h+1)} - \mathbf{Z}_i - (\mathbf{R}_i^{(h)})^\top \|_F^2 + \alpha \text{Tr}((\mathbf{W}_i^{(h+1)})^\top \mathbf{D}_{\mathbf{W}} \mathbf{W}_i^{(h+1)}) \\
&\le \| \mathbf{E}_i\mathbf{W}_i^{(h)} - \mathbf{Z}_i - (\mathbf{R}_i^{(h)})^\top \|_F^2 + \alpha \text{Tr}((\mathbf{W}_i^{(h)})^\top \mathbf{D}_{\mathbf{W}} \mathbf{W}_i^{(h)}).
\end{aligned}
\end{equation}
Substituting back the definition of the relaxation, this is equivalent to:
\begin{equation}
\label{eq:W_trace_intermediate}
\begin{aligned}
&\| \mathbf{E}_i\mathbf{W}_i^{(h+1)} - \mathbf{Z}_i - (\mathbf{R}_i^{(h)})^\top \|_F^2 + \alpha \| \mathbf{W}_i^{(h+1)} \|_{2,1} - \alpha \Delta_W^{(h+1)} \\
&\le \| \mathbf{E}_i\mathbf{W}_i^{(h)} - \mathbf{Z}_i - (\mathbf{R}_i^{(h)})^\top \|_F^2 + \alpha \| \mathbf{W}_i^{(h)} \|_{2,1} - \alpha \Delta_W^{(h)},
\end{aligned}
\end{equation}
where $\Delta_W = \| \mathbf{W}_i \|_{2,1} - \sum_j \frac{\| \mathbf{w}_{j:} \|_2^2}{2 \| \mathbf{w}_{j:}^{(h)} \|_2}$ and $\mathbf{w}_{j:}$ is the $j$-th row of $\mathbf{W}_i$.

From Lemma~\ref{lemma:inequality}, we have $\Delta_W^{(h+1)} \le \Delta_W^{(h)}$. Applying this to Inequality Eq. ~\eqref{eq:W_trace_intermediate} yields:
\begin{equation}
\label{eq:convergenceW_final}
\begin{aligned}
&\| \mathbf{E}_i\mathbf{W}_i^{(h+1)} - \mathbf{Z}_i - (\mathbf{R}_i^{(h)})^\top \|_F^2 + \alpha \| \mathbf{W}_i^{(h+1)} \|_{2,1} \\
&\le \| \mathbf{E}_i\mathbf{W}_i^{(h)} - \mathbf{Z}_i - (\mathbf{R}_i^{(h)})^\top \|_F^2 + \alpha \| \mathbf{W}_i^{(h)} \|_{2,1}.
\end{aligned}
\end{equation}
This confirms that $\Phi(\mathbf{W}_i^{(h+1)}, \mathbf{R}_i^{(h)}) \le \Phi(\mathbf{W}_i^{(h)}, \mathbf{R}_i^{(h)})$.

\textbf{Part 2: Update of $\mathbf{R}_i$ (with $\mathbf{W}_i$ fixed).}
A similar argument holds for the update of $\mathbf{R}_i^{(h+1)}$ with $\mathbf{W}_i^{(h+1)}$ fixed. We similarly obtain:
\begin{equation}
\label{eq:convergenceR_final}
\begin{aligned}
&\| \mathbf{E}_i\mathbf{W}_i^{(h+1)} - \mathbf{Z}_i - (\mathbf{R}_i^{(h+1)})^\top \|_F^2 + \beta \| \mathbf{R}_i^{(h+1)} \|_{2,1} \\
&\le \| \mathbf{E}_i\mathbf{W}_i^{(h+1)} - \mathbf{Z}_i - (\mathbf{R}_i^{(h)})^\top \|_F^2 + \beta \| \mathbf{R}_i^{(h)} \|_{2,1}.
\end{aligned}
\end{equation}
This confirms that $\Phi(\mathbf{W}_i^{(h+1)}, \mathbf{R}_i^{(h+1)}) \le \Phi(\mathbf{W}_i^{(h+1)}, \mathbf{R}_i^{(h)})$.

\textbf{Combining both parts} from Eq. \eqref{eq:convergenceW_final} and Eq. \eqref{eq:convergenceR_final}, we establish the chain of inequalities:
\begin{equation}
\Phi^{(h+1)} \le \Phi(\mathbf{W}_i^{(h+1)}, \mathbf{R}_i^{(h)}) \le \Phi^{(h)},
\end{equation}
where $\Phi^{(h)} = \Phi(\mathbf{W}_i^{(h)}, \mathbf{R}_i^{(h)})$. Since the objective function $\Phi$ is bounded below by zero, the sequence $\{ \Phi^{(h)} \}_{i=1}^{50}$ converges.
\end{proof}

\subsection{Quality Regression and Score Aggregation}

The final stage of our pipeline uses the selected, informative patches for quality prediction. Let $\mathbf{E}^*_i = \{ \mathbf{e}^*_1, \mathbf{e}^*_2, \dots, \mathbf{e}^*_k \}$ represent the set of $k$ selected patch embeddings for image $\mathbf{M}_i$, obtained from the algorithm in Sec.~\ref{sec:emb_sel}.

\updated{
\subsubsection{Patch-Level Quality Prediction}
We employ a multi-layer perceptron (MLP) regressor to learn the mapping from a patch embedding to a patch-level quality score. The MLP, parameterized by $\vartheta$, is defined as:
\begin{equation}
\text{MOS}_j = f_{\text{mlp}}(\mathbf{e}^*_j; \vartheta),
\end{equation}
where $\text{MOS}_j$ is the predicted mean opinion score (MOS) for the $j$-th selected patch. The MLP consists of an input layer, a single hidden layer with ReLU activation, and a final linear output layer. Specifically, the network architecture comprises a linear layer that projects the input features to 512 dimensions, followed by a ReLU non-linearity, and a final linear layer that maps to the output logits.}

The network is trained to minimize the mean squared error (MSE) between its predictions and the ground-truth (MOS) assigned to the entire parent image. This encourages the model to learn patch-level features that are indicative of global perceptual quality. The loss function for a mini-batch of size $b$ is:
\begin{equation}
\mathcal{L}_{\text{MSE}} = \frac{1}{b} \sum_{j=1}^{b} \left( \text{MOS} - \text{MOS}_j \right)^2,
\end{equation}
where $\text{MOS}$ is the ground-truth obtained from subjective data of the image from which $\mathbf{e}^*_j$ is computed.

\updated{
\subsubsection{Image-Level Score Aggregation}
Since the model is trained on individual patches, the $k$ predicted scores $\{\text{MOS}_1, \dots, \text{MOS}_k\}$ for an image must be pooled into a single, final image-level score. We propose a robust pooling strategy that weights each patch's score based on its consensus with the others.}

Let $\mathcal{Q} = \{\text{MOS}_1, \dots, \text{MOS}_k\}$. We compute a weight $w_j$ for each score that is inversely proportional to its deviation from the median of $\mathcal{Q}$:
\begin{equation}
w_j = \frac{1}{|\text{MOS}_j - \text{median}(\mathcal{Q})| + \epsilon},
\end{equation}
where a small constant $\epsilon > 0$ is added to avoid division by zero. The final image-level quality score is then computed as the weighted average:
\begin{equation}
\overline{pMOS} = \frac{\sum_{j=1}^{k} w_j \cdot \text{MOS}_j}{\sum_{j=1}^{k} w_j}.
\end{equation}
This approach prioritizes scores that cluster around the central tendency of the predictions, reducing the influence of potential outliers and aligning with standard practices for aggregating subjective opinions~\cite{ITU-R2012}. 

\section{Experimental Results and Analysis}
\label{sec:discussion}

\updated{This section presents a comprehensive evaluation of the proposed agnostic patch selection framework. We first detail the experimental setup and then systematically analyze the results to answer key questions about our method's efficiency, robustness, and general applicability.}

\subsection{Experimental Setup}

\subsubsection{Implementation Details}
All experiments were conducted on a server equipped with an Intel Xeon Silver 4208 2.1GHz processor, 192GB of RAM, and an Nvidia Tesla V100S 32GB GPU. The proposed framework was implemented in PyTorch.

\subsubsection{Datasets}
We validate our approach on three widely-used 360-degree IQA benchmarks:
\begin{itemize}
    \item \textbf{CVIQ}~\cite{8702664}: Contains 528 distorted images from 16 references, featuring compression artifacts (JPEG, AVC, HEVC) across 11 levels. This dataset represents homogeneous, compression-focused distortions.
    \item \textbf{OIQA}~\cite{8702664}: Comprises 320 images from 16 references with diverse distortion types (JPEG, JPEG 2000, Gaussian Noise, Gaussian Blur) at 5 levels each, offering moderate heterogeneity.
    \item \textbf{MVAQD}~\cite{9334423}: Includes 300 images from 15 references with five distortion types (JPEG, JPEG 2000, HEVC, BLUR, White Noise) across 4 levels, presenting the most heterogeneous dataset.
\end{itemize}

\subsubsection{Evaluation Protocol}
We employ standard IQA evaluation metrics: Pearson linear correlation coefficient (PLCC) for prediction accuracy and Spearman's rank correlation coefficient (SRCC) for monotonicity. Following common practices, a four-parameter logistic regression is applied before computing PLCC/SRCC. To ensure statistical reliability, we perform 5-fold random train-test splits (80\%-20\%) based on reference images and report median performance across folds.

\subsection{Performance and Efficiency Analysis}

The core premise of our work is that intelligent patch selection can maintain or improve performance while significantly reducing computational load. To validate this, we analyze the minimum selection rate required to surpass the baseline performance achieved using 100\% of patches. We apply three distinct sampling strategies to generate the initial patch set $\mathcal{P}$, each with unique characteristics and perceptual biases.

\begin{figure}
    \centering
    \begin{tabular}{c}
         \includegraphics[width=\linewidth]{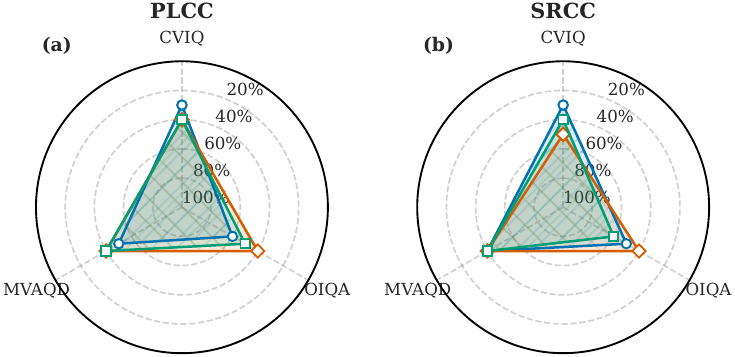}  \\
         \includegraphics[width=.6\linewidth]{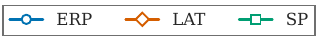}  \\
          
    \end{tabular}
    \caption{\updated{Minimum selection rates required to exceed baseline performance across different sampling strategies. Radial distance from center indicates efficiency (farther = more efficient, requiring less data).}}
    \label{fig:radae_chart}
\end{figure}


\updated{
\textbf{ERP Sampling (Equirectangular Projection)} samples patches uniformly from the equirectangular projection. Non-overlapping patches of size $128 \times 128$ are extracted from a grid over the 2D ERP image. While computationally straightforward, this method introduces significant geometric distortions, as uniform sampling in the ERP domain corresponds to non-uniform sampling on the sphere, heavily over-sampling the polar regions known to be non-significant for quality assessment. The set of patches is defined as $\mathcal{P}^{\mathrm{ERP}} = \Psi_{n}^{\mathrm{ERP}}(\mathbf{M}_i)$.}

\updated{
As shown in Fig.~\ref{fig:radae_chart}, ERP sampling demonstrates remarkable efficiency on CVIQ, requiring only 30\% of patches to exceed baseline performance. This can be attributed to ERP's uniform sampling introducing geometric distortions and redundancy, particularly near the poles. Our selection algorithm effectively filters out these problematic regions while retaining perceptually relevant content. The higher thresholds on OIQA (60\% PLCC, 50\% SRCC) and MVAQD (50\% PLCC) suggest that ERP's geometric limitations become more pronounced with diverse distortion types.}

\updated{
\textbf{LAT Sampling (Latitude-based)} addresses spherical viewing characteristics by sampling patches non-uniformly directly on the spherical domain, giving more importance to equatorial regions. Starting from the equator, patches are sampled at decreasing intervals toward the poles, with patch sizes increasing to maintain coverage. This aligns with spherical viewing patterns where users spend most time viewing content near the equator. The patch set is obtained by $\mathcal{P}^{\mathrm{LAT}} = \Psi_{n}^{\mathrm{LAT}}(\mathbf{M}_i)$, and all patches are resampled to a uniform $128 \times 128$ resolution.}

\updated{
Fig.~\ref{fig:radae_chart} reveals that LAT sampling shows exceptional consistency, achieving performance gains at 40-50\% selection rates across all datasets. This stability underscores LAT's inherent perceptual alignment with spherical viewing patterns. The method provides a robust balance between comprehensive coverage and computational efficiency, and our refinement further optimizes this already well-structured sampling by fine-tuning the selection within perceptually relevant regions.}

\updated{
\textbf{SP Sampling (Scanpath-based)} represents the most perceptually-driven approach, sampling patches centered along predicted visual scanpaths to simulate human gaze behavior. For each image, we generate multiple scanpaths using the ScanGAN360 model~\cite{9714046}, with each fixation point in a scanpath serving as the center for a patch. This ensures sampling is anchored in perceptually salient regions. The patch set is defined as $\mathcal{P}^{\mathrm{SP}} = \Psi_{n}^{\mathrm{SP}}(\mathbf{M}_i)$.}

\updated{
The results indicate that SP sampling maintains strong efficiency, requiring 40-60\% of patches across datasets. As the most perceptually-grounded strategy, SP's scanpath-based sampling inherently focuses on salient regions that are critical for quality judgment. Our refinement further optimizes this selection by removing redundancy among the already salient patches, demonstrating that even perception-driven sampling benefits from embedding-based filtering to identify the most representative samples.}

\updated{
Notably, all strategies achieve superior performance with 40-60\% fewer training samples, confirming our method's ability to discard redundant patches while preserving perceptual information critical for quality assessment. The consistent effectiveness across these fundamentally different sampling approaches underscores the robustness and general applicability of our embedding similarity-based refinement.}

\subsection{Ablation Studies and Parameter Analysis}

\subsubsection{Impact of Distance Metrics on Selection Performance}

\updated{The choice of distance metric is fundamental to our embedding similarity-based selection, as it defines the geometric space in which patch relationships are measured. We evaluate three distinct metrics, Euclidean (EUC), Manhattan (MAN), and Mahalanobis (MAH), to understand their influence on the final quality prediction performance.}

\begin{table*}[t]
\renewcommand{\arraystretch}{.8} 
\caption{\updated{Peak performance (PLCC/SRCC) achieved across all selection rates for different sampling strategies and distance metrics. Best results for each dataset-strategy combination are in \textbf{bold}.}}
\label{tab:max_performance}
\centering
\begin{tabular}{ll|cccc|cccc}
\toprule
\multirow{2}{*}{\textbf{Sampling}} & \multirow{2}{*}{\textbf{Dataset}} & \multicolumn{4}{c|}{\textbf{PLCC}} & \multicolumn{4}{c}{\textbf{SRCC}} \\
\cmidrule(lr){3-6} \cmidrule(lr){7-10}
& & Euc. & Manh. & Mah. & Avg. & Euc. & Manh. & Mah. & Avg. \\
\midrule
\multirow{3}{*}{\textbf{ERP}} 
& CVIQ   & 0.957 & 0.956 & \textbf{0.959} & 0.957 & 0.933 & 0.932 & \textbf{0.934} & 0.933 \\
& OIQA   & 0.945 & 0.952 & \textbf{0.953} & 0.950 & 0.935 & 0.946 & \textbf{0.947} & 0.943 \\
& MVAQD  & \textbf{0.918} & 0.906 & 0.918 & 0.914 & 0.877 & 0.853 & \textbf{0.881} & 0.870 \\
\midrule
\multirow{3}{*}{\textbf{LAT}} 
& CVIQ   & 0.945 & \textbf{0.945} & 0.943 & 0.944 & 0.913 & \textbf{0.916} & 0.914 & 0.915 \\
& OIQA   & \textbf{0.965} & 0.964 & 0.964 & 0.964 & 0.956 & 0.956 & \textbf{0.957} & 0.956 \\
& MVAQD  & \textbf{0.926} & 0.918 & 0.923 & 0.922 & \textbf{0.908} & 0.886 & 0.884 & 0.893 \\
\midrule
\multirow{3}{*}{\textbf{SP}} 
& CVIQ   & \textbf{0.956} & 0.952 & 0.953 & 0.954 & \textbf{0.920} & 0.920 & 0.918 & 0.919 \\
& OIQA   & 0.963 & 0.962 & \textbf{0.964} & 0.963 & \textbf{0.957} & 0.956 & 0.957 & 0.957 \\
& MVAQD  & \textbf{0.942} & 0.938 & 0.937 & 0.939 & \textbf{0.927} & 0.920 & 0.922 & 0.923 \\
\bottomrule
\end{tabular}
\end{table*}

\updated{Table~\ref{tab:max_performance} presents the peak performance achieved across all tested selection rates (10\%–90\%) for each combination of sampling strategy and distance metric. The results reveal several important patterns:}

\updated{\textbf{Mahalanobis distance} demonstrates particular strength for ERP-based sampling, achieving the best performance in 5 out of 6 cases (both PLCC and SRCC on CVIQ and OIQA, and SRCC on MVAQD). This advantage is likely because Mahalanobis distance accounts for the covariance structure of the feature space, making it better suited to handle the complex, correlated distortions introduced by equirectangular projection. By normalizing according to feature relationships, it can more effectively identify perceptually similar patches despite ERP's geometric warping.}

\updated{\textbf{Manhattan distance} shows competitive performance, particularly with LAT sampling on CVIQ where it achieves the best SRCC (0.916). Its robustness to outliers appears complementary to LAT's structured, latitude-based sampling, which already provides a perceptually coherent initial selection. The $L_1$ norm's property of being less influenced by large feature deviations makes it stable for quality assessment tasks where certain feature dimensions may not follow Gaussian distributions.}

\updated{\textbf{Euclidean distance}, while being the most computationally efficient, maintains strong performance across all configurations, achieving the best results in 4 out of 18 dataset-metric combinations. Its consistent performance underscores that the fundamental geometric relationships in the embedding space are sufficiently captured by simple $L_2$ distances for many practical applications.}

\updated{Crucially, the performance differences between metrics are insignificant, typically within 0.01–0.02 PLCC/SRCC, and often as little as 0.001–0.005. This minimal variation demonstrates that our selection framework is fundamentally robust to the specific choice of distance metric. The embedding similarity preservation mechanism appears to work effectively across different geometric interpretations of similarity, suggesting that the learned feature representations naturally cluster in ways that are detectable by multiple distance measures. This robustness has significant practical implications: it means practitioners can select distance metrics based on computational efficiency (Euclidean), robustness considerations (Manhattan), or specific data characteristics (Mahalanobis) without substantially compromising selection quality. The consistency across metrics further validates that our method captures fundamental perceptual relationships rather than being optimized for a particular distance geometry.}

\subsubsection{Impact of Embedding Dimensionality on Selection Stability}

\updated{The dimensionality reduction parameter $h$ plays a critical role in our selection algorithm, determining the complexity of the low-dimensional space where similarity relationships are preserved. To investigate its impact, we evaluated performance across a wide spectrum of $h$ values, from minimal (1) to full dimensionality (2048), including intermediate values $\{1, \dots, 10, 20, 61, 102, 204, 512, 1024, 2048\}$.}

\begin{figure*}[t]
    \centering
    \includegraphics[width=\linewidth]{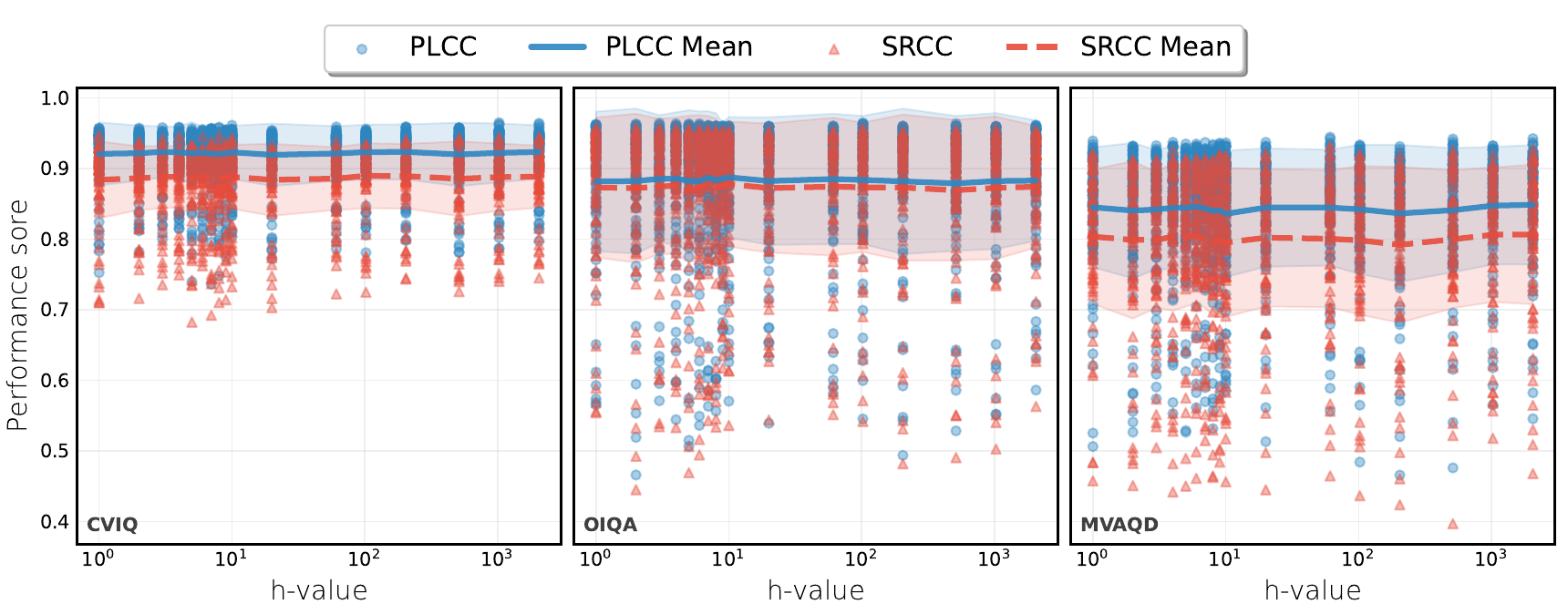}
    \caption{\updated{Performance trends of PLCC and SRCC metrics as a function of the h-parameter across three datasets (CVIQ: left, OIQA: midle, MVAQD: right). The scatter points represent individual performance measurements, while solid and dashed lines show mean trends with confidence intervals ($\pm$1 standard deviation).}}
    \label{fig:trend_h}
\end{figure*}

\updated{Fig.~\ref{fig:trend_h} presents a comprehensive analysis of how embedding dimensionality affects both the central tendency and stability of quality prediction performance. Two profound insights emerge from this investigation:}

\updated{Consistent stability in mean performance is observed across the entire dimensionality spectrum. The average PLCC and SRCC values remain largely unchanged regardless of whether we use the first components ($h \le 10$) or the complete 2048-dimensional embedding space. This finding has significant implications: it demonstrates that the most essential perceptual information for quality assessment is encoded in the dominant eigenvectors of the similarity matrix. Even with extreme dimensionality reduction, retaining as little as 0.05\% of the original feature dimensions, our selection algorithm can identify patches that are nearly as informative as those selected using the full feature space. This shows that perceptual quality relationships manifest as strong, low-rank structures in the deep feature embeddings, which can be effectively captured with minimal dimensionality.}

\updated{Besides, we observe that a dataset-dependent variance pattern emerges, which reveals fundamental differences in how distortion characteristics affect selection stability. CVIQ, with its homogeneous compression artifacts, exhibits consistently low variance across all dimensionality settings. This stability indicates that when distortions are uniform and globally consistent, patch selection operates predictably regardless of the feature subspace dimensionality. The selection process reliably identifies similar quality-relevant patterns throughout the dataset. In contrast, OIQA and MVAQD, characterized by heterogeneous distortions including blur, noise, and various compression types, show substantially higher performance variance. This variability indicates that with diverse distortions, the optimal patch selection becomes more sensitive to both the selection rate and the specific feature subspace used. Different dimensional reductions may emphasize different aspects of the complex distortion landscape, leading to greater variability. This sensitivity underscores that for heterogeneous quality assessment tasks, the interaction between selection strategy and feature representation requires more careful consideration.}

\updated{The practical implication of these findings is twofold: first, the stability of mean performance enables substantial computational savings through aggressive dimensionality reduction without sacrificing selection quality; second, the variance patterns provide valuable diagnostic information about dataset complexity and the inherent difficulty of the quality assessment task. The consistent performance across dimensionalities further validates that our similarity preservation approach captures fundamental perceptual relationships rather than relying on high-dimensional feature overfitting.}

\subsubsection{Statistical Significance Analysis of Experimental Factors}

\updated{To rigorously quantify the relative importance of each experimental factor and their interactions, we conducted a multi-factor analysis of variance (ANOVA). The latter is performed separately for each dataset (CVIQ, OIQA, MVAQD) and evaluation metric (PLCC, SRCC), with the following factors as independent variables: sampling method (ERP, LAT, SP), distance metric (EUC, MAN, MAH), log-transformed selection rate, and log-transformed embedding dimensionality ($h$). This comprehensive approach allows us to precisely determine which factors has a statistically reliable influence on the final quality prediction accuracy.}

\begin{figure}[t]
    \centering
    \setlength{\tabcolsep}{0pt}
    \begin{tabular}{c}
        \includegraphics[width=0.85\linewidth]{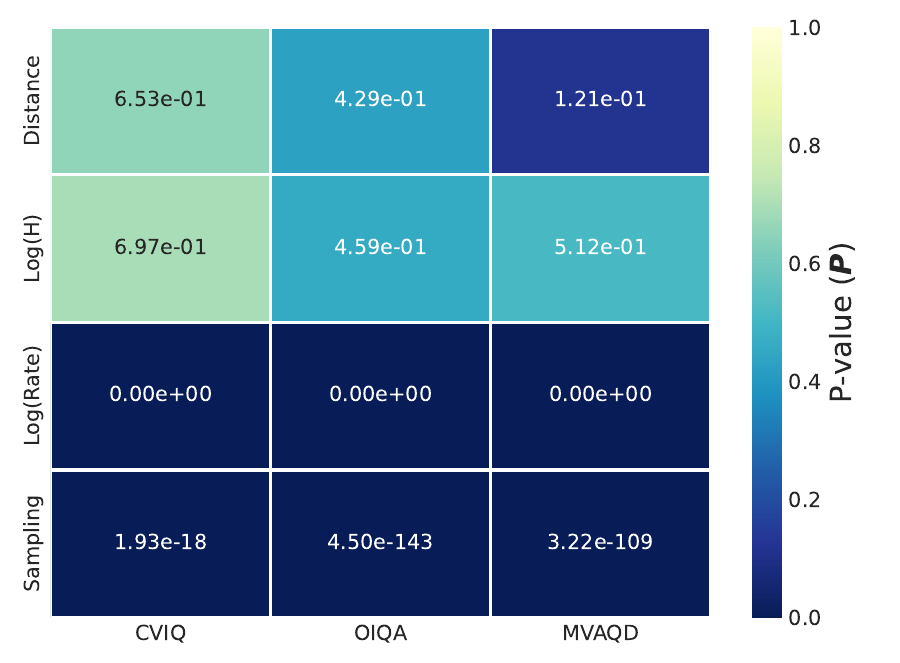}\\
        \includegraphics[width=0.85\linewidth]{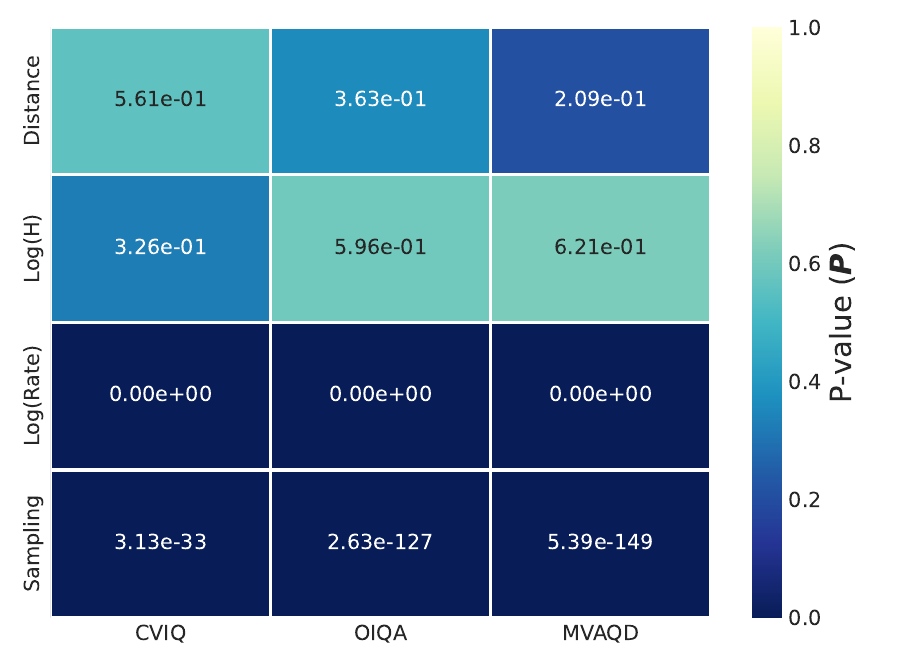}\\
    \end{tabular}
    \caption{Statistical significance of various factors (y-axis) on the performances: (top) PLCC (bottom) SRCC, across the used datasets (x-axis).}
    \label{fig:significance}
\end{figure}

\updated{Fig.~\ref{fig:significance} presents the statistical significance results as $-\log_{10}(p$-value) heatmaps, providing a clear visualization of factor importance. The analysis yields several conclusions:}

\updated{The first observation that emerges is that selection rate and sampling method are dominant factors, achieving extreme statistical significance across all three datasets and both performance metrics. The selection rate's profound influence demonstrates a strong, non-linear relationship between the quantity of selected patches and prediction stability, validating our core hypothesis that reliable data curation is crucial for effective quality assessment. Similarly, the consistent significance of the sampling method confirms that the initial spatial selection strategy fundamentally shapes the information available for quality regression, with different sampling approaches providing complementary perceptual coverage.}

\updated{Besides, the distance metric and embedding dimensionality show no statistical significance ($p \gg 0.05$ across all conditions), firmly establishing their secondary role in the selection framework. This non-significance is particularly noteworthy, it demonstrates that our embedding similarity-based selection operates effectively regardless of the specific geometric distance measure or the dimensionality of the reduced space. The algorithm's performance is driven by the fundamental similarity relationships in the feature space rather than being sensitive to the peculiarities of the distance metric. }

\updated{This clear hierarchy of factor importance has practical implications. The statistical dominance of selection rate and sampling method confirms that the focus should be oriented toward computational resources. Conversely, the robustness to distance metric and dimensionality choices significantly reduces the hyperparameter search space, making our framework more accessible and easier to deploy in practical applications. Users can confidently select computationally efficient options, for instance Euclidean distance with moderate $h \le 10$ values, without compromising selection quality.}

\subsubsection{Convergence Analysis of the Selection Algorithm}

\updated{The optimization problem defined in Eq.~\eqref{objectifve} is solved using an alternating minimization approach between the transformation matrix $\mathbf{W}_i$ and the residual matrix $\mathbf{R}_i$. To validate the stability and efficiency of this optimization strategy, we analyze the convergence behavior across different parameter configurations, including all three distance metrics and a representative subset of embedding dimensionalities ($h$).}


\begin{figure*}
    \centering
    \includegraphics[width=\linewidth]{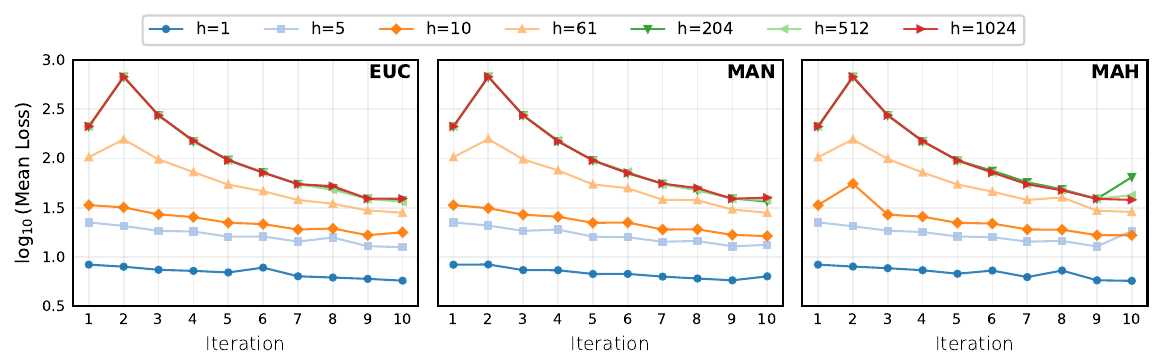}
    \caption{\updated{Convergence behavior of the objective function (log-scale) across distance metrics and a subset of embedding dimensionalities ($h$).}}
    \label{fig:conv}
\end{figure*}

\updated{Fig.~\ref{fig:conv} illustrates the convergence patterns of our optimization algorithm, revealing several important characteristics:}

\updated{Overall, we observe a consistent monotonic convergence across all experimental conditions. The objective function value decreases with each iteration, validating the effectiveness of our alternating optimization strategy. The descent indicates that the algorithm successfully navigates the optimization landscape without encountering pathological behaviors such as oscillations or divergence, which can alter non-convex problems. Besides, the observed dimensionality-dependent convergence dynamics reveal an interesting pattern: higher values of $h$, such as 204, 512, 1024, exhibit steeper initial loss reduction but require slightly more iterations to reach stability. This behavior aligns with theoretical expectations, as higher-dimensional spaces contain more parameters to optimize, leading to greater initial error but also providing more degrees of freedom for rapid initial improvement. Conversely, lower-dimensional configurations ($h \leq 64$) converge more rapidly due to their simpler optimization landscapes, though they start from lower initial loss values.}

\updated{Moreover, a robustness across distance metrics is clearly demonstrated, with all three metrics, Euclidean, Manhattan, and Mahalanobis, exhibiting nearly identical convergence patterns for equivalent $h$ values. This consistency further reinforces our earlier finding that the specific choice of distance metric has minimal impact on the selection framework's behavior. The optimization process appears equally effective regardless of the geometric properties of the distance measure used.}

\updated{Finally, the practical convergence efficiency is a key strength of our approach. All configurations achieve stable convergence within 6-8 iterations, with minimal improvement observed beyond this point. This rapid convergence has important implications for computational efficiency.}

\subsubsection{Feature Manifold Coverage and Distortion Awareness}

\updated{To rigorously ascertain that the proposed selection algorithm is distortion-aware and optimizes feature diversity according to the underlying image statistics, we analyze the normalized area coverage (NAC) saturation curves. The NAC quantitatively assesses the representativeness and spread of the selected patch subset $\mathcal{P}^{*}$ relative to the entire set of available patches $\mathcal{P}$, serving as a crucial measure of selection efficiency. A value of $\mathrm{NAC} \approx 1.0$ indicates that the selected patches sufficiently span the entire feature manifold, suggesting minimal loss of unique, high-value information.}

\begin{figure*}[h]
    \centering
    \includegraphics[width=\linewidth]{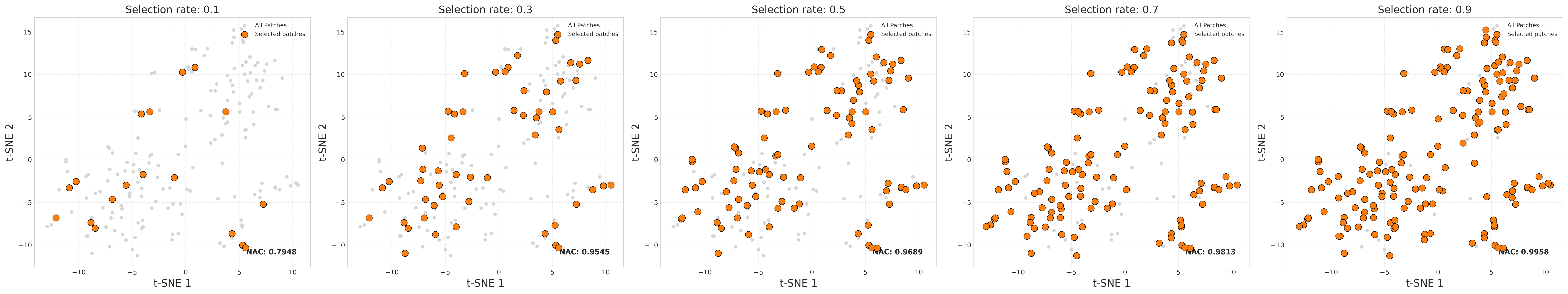}
    \caption{\updated{Visualization of the feature manifold and patch selection via t-SNE for a given 360-degree image. The complete set of patches (gray circles) defines the total feature space ($A_{\text{all}}$), while the selected patch subset (orange circles) optimally covers the manifold boundaries, defining $A_{\text{sel}}$. The efficient selection minimizes redundancy by discarding clusters of similar patches (tightly grouped gray circles) and prioritizing those that maximize the Convex Hull area.}}
    \label{fig:tsne}
\end{figure*}

\begin{figure}[ht]
    \centering
    \begin{tabular}{c}
        \includegraphics[width=.8\linewidth]{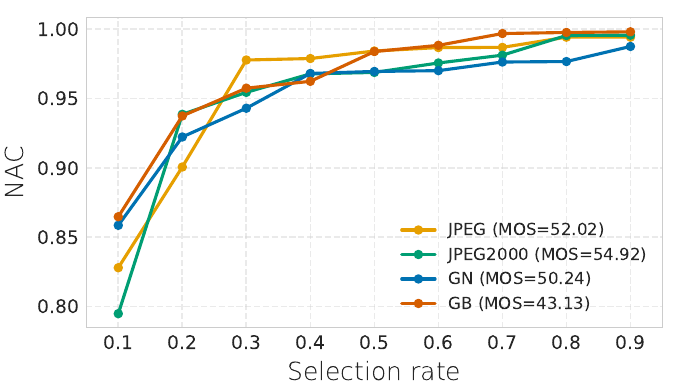}\\
        \includegraphics[width=.8\linewidth]{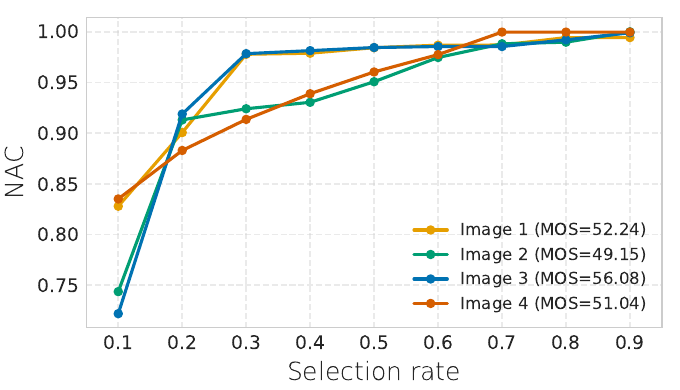}\\
    \end{tabular}
    \caption{\updated{Normalized area coverage (NAC) saturation analysis, demonstrating the selection algorithm's adaptive efficiency. (top) Distortion awareness compares the NAC curves for the same source image under four different distortion types. (bottom) Content robustness compares the NAC curves for four different images under the same JPEG distortion.}}
    \label{fig:nac}
\end{figure}

\updated{The NAC is computed by formalizing the visualization produced by dimensionality reduction techniques. Specifically, the high-dimensional feature embeddings $\mathbf{E}$ of all input patches $\mathcal{P}$ are first projected onto a two-dimensional space ($V$) using t-distributed stochastic neighbor Embedding (t-SNE)~\cite{maaten2008visualizing}. This projection visually represents the local structure of the feature manifold. The quantitative coverage is then approximated by calculating the Convex Hull area of the projected coordinates. The computation is defined as the ratio of the selected area to the total area : $\mathrm{NAC} = \frac{A_{\text{sel}}}{A_{\text{all}}}$, where $A_{\text{all}}$ is the area of the Convex Hull formed by all projected patches $V$. This establishes the normalization factor, representing the theoretical maximum feature diversity. $A_{\text{sel}}$ is the area of the Convex Hull formed exclusively by the selected patches $V^{*} \subset V$.}

\updated{For a comprehensive analysis, we compute the NAC saturation curves across all selection rates (from $0.1$ to $0.9$) in two distinct experimental setups. Scenario 1 (same image, different distortions; Fig.~\ref{fig:nac}: top) subjected the same source image to four distinct distortion types: JPEG, JPEG 2000, Gaussian Noise (GN), and Gaussian Blur (GB). This allowed for direct, controlled comparison of how the proposed selection framework adapts to structural redundancy. Scenario 2 (same distortion, different images; Fig.~\ref{fig:nac}: bottom) subjected four different source images (with varying MOS scores) to the same intermediate JPEG distortion, testing the robustness of the selection process across diverse image content and inherent complexity.}

\updated{The analysis of feature manifold saturation (Scenario 1) reveals a strong correlation between the nature of the distortion and selection efficiency. For structured/sparse distortions (JPEG, JPEG 2000), the curves exhibit the fastest saturation, reaching $\mathrm{NAC} > 0.99$ by the $0.7$ selection rate. Compression artifacts create large, highly uniform, and redundant feature clusters. Consequently, the residual analysis component of the algorithm quickly identifies the manifold boundaries (key content areas and distinct artifacts). Once these unique points are selected, subsequent selections yield minimal increase in $A_{\text{sel}}$, confirming the algorithm's high efficiency in eliminating redundancy when the feature space is sparse and highly structured. In contrast, for distributed/dense distortions (GN, GB), GN curve demonstrates the slowest saturation, never reaching $1.0$ within the tested range. This is due to the noise introducing unique perturbations into the feature vector of virtually every patch, creating a dense and continuous feature manifold where perfect redundancy is rare. The selection process must thus sample a higher proportion of patches to fully represent the subtle, unique variance, leading to sustained increases in NAC. The GB curve, while starting high, quickly plateaus, reflecting a contracted manifold where high-frequency variance is uniformly removed, allowing the algorithm to find feature redundancy quickly after the initial content boundaries are captured. Fig.~\ref{fig:tsne} illustrates the visualization of the feature manifold via t-SNE, clearly showing the distinct clusters of the unselected (redundant or irrelevant) patches relative to the selected (diverse and representative) subset.}

\updated{The robustness to image content (Scenario 2) curves demonstrate high consistency, with all four profiles following a tight, characteristic saturation pattern. Despite images having different MOSs and inherently diverse content, the NAC curves exhibit rapid convergence and consistency between the $0.3$ and $0.7$ selection rates. All curves achieve near-full coverage ($\mathrm{NAC} > 0.98$) by the $0.8$ rate, mirroring the efficient saturation observed for structured distortion in Scenario 1. The Implication of this high consistency is that the selection algorithm successfully abstracts the content of the image and focuses primarily on the distortion-induced feature redundancy and the geometry of the embedding manifold. Thus, the proposed joint optimization framework is robust to varying content, ensuring the selected subset efficiently samples the feature space regardless of the original image's complexity, provided the distortion type (JPEG) is held consistent. The minor variations observed only in the initial $0.1$ and $0.2$ rates are likely attributable to differences in the concentration of high-saliency patches across the four source images.}

\updated{In conclusion, the NAC analysis validates the efficacy of the patch selection process, confirming its ability to adapt its efficiency based on the statistical properties of the distortion (distortion-awareness) while maintaining robustness across a wide range of image content.}

\subsection{Generalization to State-of-the-Art Methods}

\updated{To demonstrate the broad applicability and practical utility of our approach, we integrated the embedding similarity-based selection as a preprocessing step with four representative state-of-the-art 360-IQA models spanning different architectural paradigms: SAP-Net~\cite{9428390} and SAL-360IQA~\cite{s23218676} (patch-based CNN), MC360IQA~\cite{8702664} (multi-channel cubemap CNN), and Assessor360~\cite{NEURIPS2023_ccf4a732} (transformer-based with multi-sequence sampling). This diverse selection tests our method's compatibility with fundamentally different quality assessment strategies. For this integration, we employed Euclidean distance and set the embedding dimensionality to $h=10$ based on our ablation studies showing robust performance with these computationally efficient hyperparameters.}

\begin{table*}[t]
\centering
\caption{\updated{Performance, in terms of PLCC/SRCC, with embedding-similarity selection applied to SOTA methods. \textbf{Bold} indicates the selection rate where performance first exceeds baseline in either PLCC or SRCC. Gray cells highlight values lower than the baseline. The $\Delta$ column shows maximum improvement over baseline.}}
\label{tab:sota_delta}
\resizebox{\textwidth}{!}{
\begin{tabular}{l|l|ccccccccc|cc}
\toprule
\textbf{Dataset} & \textbf{Method} & 0.1 & 0.2 & 0.3 & 0.4 & 0.5 & 0.6 & 0.7 & 0.8 & 0.9 & \textbf{Base} & $\mathbf{\Delta}$ \\
\midrule
\multirow{4}{*}{CVIQ} 
& SAP-Net & \cellcolor{gray!20}0.687/\cellcolor{gray!20}0.642 & \cellcolor{gray!20}0.804/\cellcolor{gray!20}0.788 & \cellcolor{gray!20}0.883/\cellcolor{gray!20}0.865 & \cellcolor{gray!20}0.934/\cellcolor{gray!20}0.917 & 0.953/\textbf{0.947} & 0.954/0.948 & \textbf{0.957}/0.950 & 0.952/0.946 & \cellcolor{gray!20}0.949/0.943 & 0.955/0.945 & 0.002/0.005 \\
& MC360IQA & \cellcolor{gray!20}0.703/\cellcolor{gray!20}0.665 & \cellcolor{gray!20}0.823/\cellcolor{gray!20}0.807 & \cellcolor{gray!20}0.903/\cellcolor{gray!20}0.892 & \cellcolor{gray!20}0.945/\cellcolor{gray!20}0.933 & \cellcolor{gray!20}0.955/0.950 & 0.962/\textbf{0.958} & \textbf{0.966}/0.960 & 0.964/0.959 & \cellcolor{gray!20}0.961/0.952 & 0.965/0.955 & 0.001/0.005 \\
& SAL-360IQA & \cellcolor{gray!20}0.728/\cellcolor{gray!20}0.684 & \cellcolor{gray!20}0.846/\cellcolor{gray!20}0.823 & \cellcolor{gray!20}0.914/\cellcolor{gray!20}0.893 & \cellcolor{gray!20}0.944/\cellcolor{gray!20}0.923 & \cellcolor{gray!20}0.950/0.945 & 0.956/\textbf{0.952} & 0.958/0.955 & \textbf{0.966}/0.957 & 0.956/0.952 & 0.960/0.950 & 0.006/0.007 \\
& Assessor360 & \cellcolor{gray!20}0.753/\cellcolor{gray!20}0.703 & \cellcolor{gray!20}0.884/\cellcolor{gray!20}0.867 & \cellcolor{gray!20}0.947/\cellcolor{gray!20}0.926 & \cellcolor{gray!20}0.973/\cellcolor{gray!20}0.964 & \cellcolor{gray!20}0.980/\cellcolor{gray!20}0.975 & \textbf{0.985}/\textbf{0.980} & 0.986/0.982 & 0.984/0.981 & \cellcolor{gray!20}0.982/0.976 & 0.984/0.980 & 0.002/0.002 \\
\midrule
\multirow{4}{*}{OIQA} 
& SAP-Net & \cellcolor{gray!20}0.653/\cellcolor{gray!20}0.624 & \cellcolor{gray!20}0.787/\cellcolor{gray!20}0.759 & \cellcolor{gray!20}0.865/\cellcolor{gray!20}0.847 & \cellcolor{gray!20}0.923/\cellcolor{gray!20}0.904 & \cellcolor{gray!20}0.948/0.940 & 0.950/\textbf{0.944} & \textbf{0.954}/0.946 & 0.950/0.944 & \cellcolor{gray!20}0.948/0.941 & 0.952/0.942 & 0.002/0.004 \\
& MC360IQA & \cellcolor{gray!20}0.686/\cellcolor{gray!20}0.648 & \cellcolor{gray!20}0.809/\cellcolor{gray!20}0.784 & \cellcolor{gray!20}0.889/\cellcolor{gray!20}0.858 & \cellcolor{gray!20}0.934/\cellcolor{gray!20}0.924 & \cellcolor{gray!20}0.950/0.945 & 0.957/\textbf{0.952} & \textbf{0.964}/0.955 & 0.958/0.953 & \cellcolor{gray!20}0.954/0.948 & 0.960/0.952 & 0.004/0.003 \\
& SAL-360IQA & \cellcolor{gray!20}0.707/\cellcolor{gray!20}0.668 & \cellcolor{gray!20}0.824/\cellcolor{gray!20}0.803 & \cellcolor{gray!20}0.885/\cellcolor{gray!20}0.883 & \cellcolor{gray!20}0.933/\cellcolor{gray!20}0.921 & \cellcolor{gray!20}0.950/0.945 & 0.955/\textbf{0.950} & \textbf{0.958}/0.953 & 0.960/0.955 & \cellcolor{gray!20}0.956/0.950 & 0.958/0.948 & 0.002/0.007 \\
& Assessor360 & \cellcolor{gray!20}0.728/\cellcolor{gray!20}0.685 & \cellcolor{gray!20}0.857/\cellcolor{gray!20}0.829 & \cellcolor{gray!20}0.903/0.903 & \cellcolor{gray!20}0.952/0.943 & \cellcolor{gray!20}0.965/0.960 & \cellcolor{gray!20}0.970/0.965 & \textbf{0.976}/\textbf{0.967} & \cellcolor{gray!20}0.971/0.965 &\cellcolor{gray!20} 0.969/0.963 & 0.972/0.966 & 0.004/0.001 \\
\midrule
\multirow{4}{*}{MVAQD} 
& SAP-Net & \cellcolor{gray!20}0.604/\cellcolor{gray!20}0.568 & \cellcolor{gray!20}0.723/\cellcolor{gray!20}0.689 & \cellcolor{gray!20}0.827/\cellcolor{gray!20}0.807 & \cellcolor{gray!20}0.883/\cellcolor{gray!20}0.863 & 0.938/\textbf{0.930} & \textbf{0.940}/0.932 & 0.945/0.932 & 0.936/0.929 & \cellcolor{gray!20}0.931/0.924 & 0.940/0.925 & 0.005/0.007 \\
& MC360IQA & \cellcolor{gray!20}0.629/\cellcolor{gray!20}0.586 & \cellcolor{gray!20}0.741/\cellcolor{gray!20}0.708 & \cellcolor{gray!20}0.835/\cellcolor{gray!20}0.825 & \cellcolor{gray!20}0.902/\cellcolor{gray!20}0.879 & 0.945/\textbf{0.940} & \textbf{0.950}/0.945 & 0.951/0.943 & 0.946/0.940 & \cellcolor{gray!20}0.941/0.936 & 0.948/0.935 & 0.003/0.008 \\
& SAL-360IQA & \cellcolor{gray!20}0.647/\cellcolor{gray!20}0.607 & \cellcolor{gray!20}0.763/\cellcolor{gray!20}0.725 & \cellcolor{gray!20}0.853/\cellcolor{gray!20}0.833 & \cellcolor{gray!20}0.903/\cellcolor{gray!20}0.879 & 0.940/\textbf{0.935} & \textbf{0.945}/0.940 & 0.948/0.943 & 0.950/0.945 & 0.946/0.940 & 0.945/0.933 & 0.005/0.012 \\
& Assessor360 & \cellcolor{gray!20}0.685/\cellcolor{gray!20}0.643 & \cellcolor{gray!20}0.826/\cellcolor{gray!20}0.789 & \cellcolor{gray!20}0.893/\cellcolor{gray!20}0.852 & \cellcolor{gray!20}0.934/\cellcolor{gray!20}0.913 & \cellcolor{gray!20}0.960/0.955 & \cellcolor{gray!20}0.965/0.960 & 0.968/\textbf{0.963} & \cellcolor{gray!20}0.966/0.961 & \cellcolor{gray!20}0.962/0.956 & 0.969/0.962 & -0.001/0.001 \\
\bottomrule
\end{tabular}}
\end{table*}

\updated{Table~\ref{tab:sota_delta} presents a comprehensive evaluation across selection rates from 10\% to 90\%, revealing several insights that demonstrate the universal value of our approach:}

\updated{From the performances, we can see that all methods reach or exceed their baseline performance using only 60-70\% of the original training data, with optimal performance typically achieved at 60-80\% selection rates rather than the 30-50\% rates observed in our core validation with ResNet-50. This pattern can be attributed to the fact that these SOTA models are specifically designed and optimized for 360-degree IQA, already incorporating sophisticated mechanisms for content selection and quality assessment. Consequently, their baseline performance is already highly efficient, leaving less room for considerable improvements compared to using a generic feature extractor like ResNet-50. Despite this, our method still achieves a substantial 20-40\% reduction in computational requirements for both training and inference while maintaining or enhancing the perceptual quality prediction accuracy.}

\updated{Additionally, architectural independence is clearly demonstrated through the consistent benefits across fundamentally different model paradigms. The method works equally well with:
\begin{itemize}
    \item \textbf{Multi-channel CNNs} (MC360IQA) where our selection reduces the computational burden of processing multiple cubemap faces.
    \item \textbf{Patch-based networks} (SAP-Net and SAL-360IQA) where it optimizes the initial patch sampling.
    \item \textbf{Transformer architectures} (Assessor360) where it enhances the multi-sequence sampling strategy.
\end{itemize}}

\updated{This architectural agnosticism underscores that the benefits of intelligent sample selection transcend specific model designs and are fundamental to data-efficient quality assessment.} 

\updated{Overall, practical deployment advantages emerge from these findings. Our embedding similarity selection can be seamlessly integrated as a refinement module into existing 360-IQA pipelines without requiring architectural changes. The method provides an immediate pathway to reduce computational costs, accelerate training times, and potentially improve model generalization across all major classes of contemporary IQA architectures.}

\section{Conclusion}
\label{sec:conc}

\updated{This paper has introduced a novel embedding similarity-based framework for refining patch selections in 360-degree IQA. We have addressed a critical but often overlooked aspect of the pipeline: the redundancy in the data \textit{"after"} it has been sampled. Our core contribution is an adaptive selection algorithm that can be applied after any initial sampling strategy to distill a large, potentially redundant set of patches into a compact, highly informative subset. This is formalized as an optimization problem that preserves perceptual similarity in a low-dimensional embedding space while using robust residual analysis to discard outliers.}

\updated{The extensive validation confirms the robustness of the proposed approach. When applied to standard sampling methods, our refinement maintains high accuracy using 40-50\% less data. Furthermore, its robustness to parameters like distance metrics and embedding dimensionality makes it practical and easy to deploy. Most significantly, its value as a generic, model-agnostic module was proven by integrating it with diverse state-of-the-art IQA models, consistently reducing their computational load by 20-40\% without sacrificing performance.}

\updated{This work establishes that a dedicated data refinement step, separate from initial sampling and model architecture, is a highly effective lever for improving efficiency in 360-degree IQA. By providing a universal method to curate higher-quality training and inference data, we offer a practical pathway to more scalable immersive media assessment. Future work will focus on making this refinement end-to-end trainable and extending its principles to 360-degree video.}

\bibliographystyle{IEEEtran}
\bibliography{main}

@article{residual_analysis, 
  title={Outlier detection using nonconvex penalized regression},
  author={Y. She and AB. Owen},
  journal={Journal of the American Statistical Association},
  volume={106},
  number={494},
  pages={626--639},
  year={2011},
  publisher={Taylor \& Francis}
}

@article{similarity,
  author={Z. Zhao and L. Wang and H. Liu and J. Ye},
  journal={IEEE TKDE}, 
  title={On Similarity Preserving Feature Selection}, 
  year={2013},
  volume={25},
  number={3},
  pages={619-632},
  doi={10.1109/TKDE.2011.222}}

@article{mahalanobis2cv,
  title={On the generalized distance in statistics},
  author={Mahalanobis, P},
  journal={Proc. of the Nation. Acad. Sci.,(India)},
  year={1936},
  volume={2},
  pages={49--5}
}

@inproceedings{l21-norms-paper,
    author = {F. Nie and H. Huang and X. Cai and C. Ding},
    title = {Efficient and robust feature selection via joint $\ell_{2,1}$-norms minimization},
    year = {2010},
    publisher = {Curran Associates Inc.},
    booktitle = {NIPS},
    pages = {1813–1821},
    address = "Vancouver, Canada"
}

@article{yan2025max360iq,
  title={{Max360IQ}: Blind omnidirectional image quality assessment with multi-axis attention},
  author={J. Yan and Z. Tan and Y. Fang and J. Rao and Y. Zuo},
  journal={Pattern Recognition},
  pages={111429},
  year={2025},
  publisher={Elsevier}
}

@book{bishop2006pattern,
  title={Pattern recognition and machine learning},
  author={CM. Bishop and NM. Nasrabadi},
  volume={4},
  number={4},
  year={2006},
  publisher={Springer}
}

@article{fan2024omiqnet,
  title={OmiQnet: Multiscale feature aggregation convolutional neural network for omnidirectional image assessment},
  author={Fan, Yu and Chen, Chunyi},
  journal={Applied Intelligence},
  pages={1--17},
  year={2024},
  publisher={Springer}
}

@ARTICLE{10297422,
  author={Liu, Lixiong and Ma, Pingchuan and Wang, Chongwen and Xu, Dong},
  journal={IEEE SPL}, 
  title={Omnidirectional Image Quality Assessment With Knowledge Distillation}, 
  year={2023},
  volume={30},
  number={},
  pages={1562-1566},
  doi={10.1109/LSP.2023.3327908}}

@misc{360_growth,
  author       = {Market Research Intellect},
  title        = {360-Degree Camera Market Poised for Significant Expansion: Projected to Grow from USD 3.2 Billion in 2024 to USD 6.8 Billion by 2031},
  year         = {2024},
  url          = {https://shorturl.at/6fI8H},
  note         = {Accessed: 2024-09-24}
}

@INPROCEEDINGS{8486584,
  author={S. Chen and Y. Zhang and Y. Li and Z. Chen and Z. Wang},
  booktitle={IEEE ICME}, 
  title={Spherical Structural Similarity Index for Objective Omnidirectional Video Quality Assessment}, 
  year={2018},
  volume={},
  number={},
  pages={1-6},
  address="San Diego, CA, USA"}

@article{tofighi2024omnidirectional,
  title={Omnidirectional image quality assessment with local--global vision transformers},
  author={Tofighi, Nafiseh Jabbari and Elfkir, Mohamed Hedi and Imamoglu, Nevrez and Ozcinar, Cagri and Erdem, Aykut and Erdem, Erkut},
  journal={Image and Vision Computing},
  volume={148},
  pages={105151},
  year={2024},
  publisher={Elsevier}
}

@book{garcia2015data,
  title={Data preprocessing in data mining},
  author={S. Garc{\'\i}a and J. Luengo and F. Herrera},
  volume={72},
  year={2015},
  publisher={Springer}
}

@article{perkis2020qualinet,
  title={QUALINET white paper on definitions of immersive media experience (IMEx)},
  author={A. Perkis and C. Timmerer and S. Barakovi{\'c} and J. Husi{\'c} and \textit{et al.}},
  year={2020},
  journal={ENQEMSS, 14th QUALINET meeting (online)}
}

@incollection{ALAIN20233,
    title = {Introduction to immersive video technologies},
    booktitle = {Immersive Video Technologies},
    publisher = {Academic Press},
    pages = {3-24},
    year = {2023},
    isbn = {978-0-323-91755-1},
    doi = {https://doi.org/10.1016/B978-0-32-391755-1.00007-9},
    author = {A. Martin and Z. Emin and O. Cagri and V. Giuseppe}
}

@article{zhou2023perception,
  title={Perception-Oriented U-Shaped Transformer Network for 360-Degree No-Reference Image Quality Assessment},
  author={M. Zhou and L. Chen and X. Wei and X. Liao and Q. Mao and H. Wang and H. Pu and J. Luo and T. Xiang and B. Fang},
  journal={IEEE TB},
  year={2023},
  volume={69},
  number={2},
  pages={396-405}
}

@inproceedings{yang2022tvformer,
  title={{TVFormer}: Trajectory-guided visual quality assessment on 360° images with transformers},
  author={L. Yang and M. Xu and T. Liu and L. Huo and X. Gao},
  booktitle={30th ACM ICM},
  pages={799--808},
  year={2022}
}

@ARTICLE{9964240,
  author={C. Tian and F. Shao and X. Chai and Q. Jiang and Xu, Long and Ho, Yo-Sung},
  journal={IEEE TCSVT}, 
  title={Viewport-Sphere-Branch Network for Blind Quality Assessment of Stitched 360° Omnidirectional Images}, 
  year={2023},
  volume={33},
  number={6},
  pages={2546-2560}
}

@inproceedings{8702664,
  author={W. Sun and X. Min and G. Zhai and K. Gu and H. Duan and S. Ma},
  booktitle={IEEE JSTSP}, 
  title={{MC360IQA}: A Multi-channel {CNN} for Blind 360-Degree Image Quality Assessment},
  year={2020},
  volume={14},
  number={1},
  pages={64-77},
  doi={10.1109/JSTSP.2019.2955024}
}

@inproceedings{8953510,
  author    = {C. {Li} and M. {Xu} and L. {Jiang} and S. {Zhang} and X. {Tao}},
  booktitle = {IEEE CVPR},
  title     = {Viewport Proposal CNN for 360° Video Quality Assessment},
  pages     = {10169-10178},
  doi       = {10.1109/CVPR.2019.01042}
}

@article{maaten2008visualizing,
  title={Visualizing data using t-SNE},
  author={Maaten, Laurens van der and Hinton, Geoffrey},
  journal={Journal of machine learning research},
  volume={9},
  number={Nov},
  pages={2579--2605},
  year={2008}
}

@ARTICLE{9163077,
  author={J. Xu and W. Zhou and Z. Chen},
  journal={IEEE TCSVT}, 
  title={Blind Omnidirectional Image Quality Assessment With Viewport Oriented Graph Convolutional Networks}, 
  year={2021},
  volume={31},
  number={5},
  pages={1724-1737}
}

@article{yan2025omnidirectional,
  title={Omnidirectional image quality captioning: A large-scale database and a new model},
  author={Yan, Jiebin and Tan, Ziwen and Fang, Yuming and Chen, Junjie and Jiang, Wenhui and Wang, Zhou},
  journal={IEEE Transactions on Image Processing},
  year={2025},
  publisher={IEEE}
}

@article{sendjasni2023pw,
  title={{PW-360IQA}: Perceptually-weighted multichannel cnn for blind 360-degree image quality assessment},
  author={Sendjasni, Abderrezzaq and Larabi, Mohamed-Chaker},
  journal={Sensors},
  volume={23},
  number={9},
  pages={4242},
  year={2023},
  publisher={MDPI}
}

@Article{s23218676,
AUTHOR = {A. Sendjasni and M-C. Larabi},
TITLE = {Attention-Aware Patch-Based CNN for Blind 360-Degree Image Quality Assessment},
JOURNAL = {Sensors},
VOLUME = {23},
YEAR = {2023},
NUMBER = {21},
ARTICLE-NUMBER = {8676}
}

@ARTICLE{9334423,
  author={H. Jiang and G. Jiang and M. Yu and Y. Zhang and Y. Yang and Z. Peng and F. Chen and Q. Zhang},
  journal={IEEE TIP}, 
  title={Cubemap-Based Perception-Driven Blind Quality Assessment for 360-degree Images}, 
  year={2021},
  volume={30},
  number={},
  pages={2364-2377}
}

@INPROCEEDINGS{9428390,
  author={L. Yang and M. Xu and X. Deng and B. Feng},
  booktitle={IEEE ICME}, 
  title={Spatial Attention-Based Non-Reference Perceptual Quality Prediction Network for Omnidirectional Images}, 
  year={2021},
  pages={1-6},
  address="Shenzhen, China"
}

@article{8638985,
  author  = {HG. Kim and H. Lim and YM. Ro},
  journal = {IEEE TCSVT},
  title   = {Deep Virtual Reality Image Quality Assessment With Human Perception Guider for Omnidirectional Image},
  year    = {2020},
  volume  = {30},
  number  = {4},
  pages   = {917-928},
  doi     = {10.1109/TCSVT.2019.2898732}
}

@INPROCEEDINGS{10096750,
  author={NJ. Tofighiand and EM. Hedi and N. Imamoglu and C. Ozcinar and E. Erdem and A. Erdem},
  booktitle={IEEE ICASSP}, 
  title={{ST360IQ}: No-Reference Omnidirectional Image Quality Assessment With Spherical Vision Transformers}, 
  year={2023},
  volume={},
  number={},
  pages={1-5},
  address="Rhodes Island, Greece"
}

@article{liu2023dual,
  title={Dual-Level Blind Omnidirectional Image Quality Assessment Network Based on Human Visual Perception},
  author={D. Liu and L. Zhang and L. Wan and Y. Yao and J. Ma and Y. Zhang},
  journal={IJACSA},
  volume={14},
  number={9},
  year={2023}
}

@inproceedings{NEURIPS2023_ccf4a732,
author = {Wu, Tianhe and Shi, Shuwei and Cai, Haoming and Cao, Mingdeng and Xiao, Jing and Zheng, Yinqiang and Yang, Yujiu},
title = {Assessor360: multi-sequence network for blind omnidirectional image quality assessment},
year = {2023},
address = {Red Hook, NY, USA},
booktitle = {ICNIPS},
articleno = {2834},
numpages = {14},
location = {New Orleans, LA, USA},
series = {NIPS '23}
}

@inproceedings{resnet,
  title     = {Deep residual learning for image recognition},
  author    = {K. He and X. Zhang and S. Ren and J. Sun},
  booktitle = {IEEE CVPR},
  pages     = {770--778},
  year      = {2016},
  address   = {Las Vegas, NV, USA}
}

@INPROCEEDINGS{9506044,
  author={A. Sendjasni and M-C. Larabi and FA. Cheikh},
  booktitle={IEEE ICIP}, 
  title={Perceptually-Weighted {CNN} For 360-Degree Image Quality Assessment Using Visual Scan-Path And {JND}}, 
  year={2021},
  volume={},
  number={},
  pages={1439-1443},
  address="Anchorage, AK, USA"
}

@ARTICLE{9432940,
  author={Y. Zhou and Y. Sun and L. Li and K. Gu and Y. Fang},
  journal={IEEE TCSVT}, 
  title={Omnidirectional Image Quality Assessment by Distortion Discrimination Assisted Multi-Stream Network}, 
  year={2021},
  volume={},
  number={},
  pages={1-1}}

@ARTICLE{9791414,
  author={A. Sendjasni and M-C. Larabi and FA. Cheikh},
  journal={IEEE TCSVT}, 
  title={Convolutional Neural Networks for Omnidirectional Image Quality Assessment: A Benchmark}, 
  year={2022},
  volume={32},
  number={11},
  pages={7301-7316}
}

@ARTICLE{8434258,
  author={M. Huang and Q. Shen and A. Ma and AC. Bovik  and P. Gupta and R. Zhou and X. Cao},
  journal={IEEE TIP}, 
  title={Modeling the Perceptual Quality of Immersive Images Rendered on Head Mounted Displays: Resolution and Compression}, 
  year={2018},
  volume={27},
  number={12},
  pages={6039-6050}
}

@book{ITU-R2012,
author = {ITU-R},
booktitle = {International Telecommunication Union},
isbn = {0920122019},
title = {{Methodology for the subjective assessment of the quality of television pictures BT Series Broadcasting service}},
volume = {13},
year = {2012}
}

@article{Sun2017,
  title   = {Weighted-to-Spherically-Uniform Quality Evaluation for Omnidirectional Video},
  author  = {Yule Sun and Ang Lu and L. Yu},
  journal = {IEEE Signal Processing Letters},
  year    = {2017},
  volume  = {24},
  pages   = {1408-1412}
}

@ARTICLE{9714046,
  author={D. Martin and A. Serrano and A. Bergman and G. Wetzstein and B. Masia},
  journal={IEEE TVCG}, 
  title={ScanGAN360: A Generative Model of Realistic Scanpaths for 360° Images}, 
  year={2022},
  volume={28},
  number={5},
  pages={2003-2013},
  doi={10.1109/TVCG.2022.3150502}}

@article{kuncheva2019instance,
  title={Instance selection improves geometric mean accuracy: a study on imbalanced data classification},
  author={L. Kuncheva and A. Arnaiz and F. D{\'\i}ez-Pastor and A. Gunn},
  journal={Progress in Artificial Intelligence},
  volume={8},
  pages={215--228},
  year={2019},
  publisher={Springer}
}

@article{chen2022deep,
  title={Deep learning for instance retrieval: A survey},
  author={W. Chen and Y. Liu and W. Wang and E. Bakker and T. Georgiou and P. Fieguth and L. Liu and MS. Lew},
  journal={IEEE TPAMI},
  volume={45},
  number={6},
  pages={7270--7292},
  year={2022}
}

\end{document}